\documentclass[lettersize,journal]{IEEEtran}
\usepackage{amsmath,amsfonts}
\usepackage{algorithmic}
\usepackage{algorithm}
\usepackage{array}
\usepackage[caption=false,font=normalsize,labelfont=sf,textfont=sf]{subfig}
\usepackage{textcomp}
\usepackage{stfloats}
\usepackage{url}
\usepackage{verbatim}
\usepackage{graphicx}
\usepackage{cite}
\usepackage{xcolor}
\usepackage{amsthm}
\usepackage{amssymb}
\usepackage{amsmath,lipsum}
\usepackage{epstopdf}
\usepackage{bm}
\usepackage{cuted}
\DeclareMathOperator*{\argmin}{argmin}

\begin{document}

\title{\huge Performance Analysis for Resource Constrained Decentralized Federated Learning Over Wireless Networks}

\author{Zhigang Yan and Dong Li,~\IEEEmembership{Senior Member, IEEE}
	\vspace{-10mm}
	\thanks{Zhigang Yan and Dong Li are with the School of Computer
		Science and Engineering, Macau University of Science and Technology, Macau, China. (e-mail: 3220005784@student.must.edu.mo and dli@must.edu.mo).
		
	}
	
}

\maketitle

\begin{abstract}
Federated learning (FL) can generate huge communication overhead for the central server, which may cause operational challenges. Furthermore, the central server's failure or compromise may result in a breakdown of the entire system. To mitigate this issue, decentralized federated learning (DFL) has been proposed as a more resilient framework that does not rely on a central server, as demonstrated in previous works. DFL involves the exchange of parameters between each device through a wireless network. To optimize the communication efficiency of the DFL system, various transmission schemes have been proposed and investigated. However, the limited communication resources present a significant challenge for these schemes. Therefore, to explore the impact of constrained resources, such as computation and communication costs on DFL, this study analyzes the model performance of resource-constrained DFL using different communication schemes (digital and analog) over wireless networks. 
Specifically, we provide convergence bounds for both digital and analog transmission approaches, enabling analysis of the model performance trained on DFL. Furthermore, for digital transmission, we investigate and analyze resource allocation between computation and communication and convergence rates, obtaining its communication complexity and the minimum probability of correction communication required for convergence guarantee. For analog transmission, we discuss the impact of channel fading and noise on the model performance and the maximum errors accumulation with convergence guarantee over fading channels. Finally, we conduct numerical simulations to evaluate the performance and convergence rate of convolutional neural networks (CNNs) and Vision Transformer (ViT) trained in the DFL framework on fashion-MNIST and CIFAR-10 datasets. Our simulation results validate our analysis and discussion, revealing how to improve performance by optimizing system parameters under different communication conditions.
\end{abstract}

\vspace{-3mm}
\section{Introduction}
\IEEEPARstart{W}{ith} the rise of the Internet of Things (IoT), there is an explosion of data generated by IoT devices \cite{ye, ye2}, which can be used to train large-scale machine learning models. However, due to communication cost and privacy concerns, it is impractical to upload all the data to the cloud or a central base station (BS) for model training \cite{R.A.}. Federated learning (FL) was proposed as a machine learning model training framework that addresses these issues, by enabling edge devices to train local models and transmit the model updates to the BS for aggregation without the need for the raw data \cite{SURVEY,MMB,avg1,avg2}. However, the limited communication resources of the BS can become a bottleneck for the communication efficiency of the FL system. To address this issue, several resource allocation strategies \cite{minE,minTE,minF,minF2}, user scheduling strategies \cite{sch1,sch2,sch3}, parameter quantization algorithms \cite{q1,q2,q3}, and advanced aggregation algorithms \cite{agg1,agg2,agg3} have been proposed. However, these strategies rely on a centralized BS, which makes the FL system vulnerable to failures or attacks. Decentralized federated learning (DFL) has been proposed as a more robust and scalable alternative, where the model training is done in a decentralized manner, with local model updates exchanged directly between devices, without the need for a central server. DFL is attracting increasing attention and research effort as it can overcome the limitations of traditional FL and offer better scalability and robustness.

DFL is rooted in the field of distributed optimization, which has been extensively studied in the past years \cite{dopt}. Classic optimization algorithms, such as gradient and subgradient descent \cite{dgd1,dgd2} and the alternating direction method of multipliers (ADMM) \cite{admm}, have been applied to distributed optimization problems, including those encountered in DFL. In distributed optimization algorithms, information exchange between nodes is usually based on randomized gossip algorithms, which have been proposed as a method for efficient and scalable information exchange in large-scale networks \cite{gossip}. The randomized gossip algorithm is a type of decentralized protocol that enables nodes to exchange information in a peer-to-peer fashion, with each node communicating with its neighbors at each iteration. This algorithm has been applied to various distributed optimization problems and is well-suited for DFL, where communication between devices is often limited and robustness concerns may require a decentralized approach. The application of distributed stochastic gradient descent (D-SGD) to DFL has been widely explored in \cite{gl1,gl2,gl3,gl4,con1,con2,con3,con4,noise1,noise2,digit1,digit2,error,dfl1,dfl2}, with several studies focused on three main areas: 1) DFL with perfect communication, 2) Application of DFL in wireless networks, and 3) Resource allocation in DFL. However, these problems are not independent of each other and are often interrelated. While several studies have experimentally verified the convergence and performance of DFL using D-SGD, as reported in \cite{gl1,gl2,gl3,gl4}, the theoretical analysis of DFL is still limited. For instance, while some works have explored the convergence of DFL, as reported in \cite{con1,con2,con3,con4}, these studies did not take into account the practical communication constraints associated with different communication (digital or analog) used in DFL. For completeness, it is worth noting that some recent studies have investigated the convergence of DFL under analog \cite{noise1,noise2} and digital communication approaches \cite{digit1,digit2}, but these studies did not address the resource-constrained problem of DFL in practical communication scenarios. For example, the transportation layer communication protocol of digital transmission used in these works is transmission control protocol (TCP), which leads to significant communication overhead due to its reliability guarantees and retransmission mechanism. To address this issue, some recent works \cite{error} have explored DFL with user datagram protocol (UDP) to reduce the communication overhead and improve the scalability. However, UDP sacrifices some reliability guarantees, which can lead to package errors in inter-device communication, and these works discuss the impact of such errors on DFL. However, these studies did not address the resource allocation problem of DFL in resource-constrained scenarios. Specifically, the computation and communication resources (e.g., bandwidth and energy) available to each device are often limited, and optimizing the allocation of these resources remains a significant challenge in DFL research. Due to the limited communication resources, the DFL algorithm that alternates between a local update and a communication incurs additional communication costs. Furthermore, a DFL algorithm that achieves a balance between the computation and communication was proposed and applied in \cite{dfl1,dfl2}. The DFL with balanced communication and computing costs over digital transmission was proposed in \cite{dfl1} with convergence analysis. Based on this algorithm, a scheduling scheme to improve the communication efficiency was proposed and verified experimentally in \cite{dfl2}. However, these works ignore the package errors caused by UDP in digital transmission with balanced recourse allocation. In summary, although DFL has been proposed and proved to converge in error-free case \cite{gl1,gl2,gl3,gl4,con1,con2,con3,con4}, compared with FL, DFL has not received enough attention. Besides, DFL under specific transmission schemes \cite{noise1,noise2,digit1,digit2} do not take into account the limited resource overhead of wireless networks, which, however, has already attracted substantial research attention in FL. Moreover, DFL that strikes a balance computing and communication resources has been paid attention to by \cite{dfl1,dfl2}, and the impact of specific communication schemes and parameters on DFL performance and specific allocation strategies has not yet been exploited.

Motivated by the above observations, in this paper, we propose a comprehensive performance and convergence analysis, which jointly considers the different communication approaches (digital and analog), resource constraints, the channel fading, noise and package errors that DFL with UDP encounters in wireless networks, and discusses the effect of the system parameters, communication conditions and connection relationship between devices. Our main contributions are summarized as follows:

\begin{itemize}

	\item{By jointly considering the digital and analog transmission approaches in DFL over wireless networks, we derive two upper bounds of the gap between the loss function of DFL and its globally optimal one in these two cases. By taking into account the practical limitations of communication resources, the probability of package errors caused by UDP in digital transmission and error accumulation in analog transmission, the derived upper bounds can provide valuable insights into the convergence and performance of DFL in digital and analog transmission scenarios. Moreover, the effect of various factors, such as the number of local training rounds, communication rounds and the number of devices are explored and analyzed.}
	
	\item{In order to gain more insights, we conduct further studies into resource allocation, convergence rates and communication complexity of DFL with digital communication. Firstly, we develop a resource allocation strategy considering computation and communication of DFL that minimizes the derived upper bound while satisfying communication and training resource constraints, achieving a balance between local training and communication rounds. Secondly, we derive the minimum number of update rounds required for DFL to achieve a predefined accuracy and analyzed the influence of system parameters, communication conditions, and connection density of devices (graph topology) on the convergence rate and communication complexity. We also investigate the convergence rates of DFL with digital transmission under different graph topologies. Besides, the minimum correct probability of communication that DFL with UDP over digital transmission can tolerate is also proposed. In addition, we analyze the impact of fading channel and noise in analog communication and explore the influence of different variables on error accumulation. Finally, we obtain the maximum error accumulation caused by channel fading and noise} that a DFL system can tolerate.
	
	\item{We perform numerical experiments to validate our analysis and investigate the resource allocation and convergence rate of DFL with digital and analog communication. Specifically, we train convolutional neural networks (CNNs) and Vision Transformer (ViT) on the fashion-MNIST and CIFAR-10 datasets in the DFL framework. The simulation results not only verify our analysis and discussions but also provide meaningful insights for system parameter settings to improve model performance. These include setting a larger number of local training rounds in resource-constrained DFL with more devices can improve performance, and selecting fewer devices to participate in DFL is beneficial when dealing with higher levels of noise power in the communication channel. Furthermore, the comparison between the proposed scheme and different benchmarks is also shown.} 
	
\end{itemize}

The remainder of this paper is organized as follows. In Section II, we introduce the system model of DFL. The performance analysis of DFL with digital communication approach is presented in Section III. The performance analysis of DFL with analog communication approach is presented in Section IV. The simulation results are shown in Section V and the concluding remarks are given in Section VI.

\section{System Model and Preliminaries}

In this paper, the DFL model under consideration consists of $N$ devices with their local data sets $\mathcal{D}_{1}, \cdots, \mathcal{D}_{N}$. We assume that the connections between each device in the DFL system form an undirected graph. This graph can be written as $\mathcal{G}(\mathcal{V},\mathcal{E})$, where $\mathcal{V}=\{1,\cdots,N\}$ is the set of nodes in graph and $\mathcal{E}\subseteq\{(i,j)\in\mathcal{V} \times \mathcal{V}~|~i\neq j \}$ denoting the set of edges, as illustrated in Fig. 1 for example.

The primary objective of DFL is to minimize the global loss function in a distributed manner without any central node. Unlike the centralized approach to model training, DFL divides the complete dataset into $N$ shares and distributes them among $N$ different nodes. In addition to parameter updating using the gradient descent (GD) on the local dataset, each node exchanges parameters with its neighboring nodes to expand the feasible domain of local parameters, ultimately achieving the optimum based on the complete dataset. Therefore, the DFL system minimizes the global loss function by carrying out multiple rounds of local updates and communication between nodes. Specifically, the global loss function at the $t$-th update round is defined as the average of each device's local loss function \cite{con1,con2,con3,con4,noise1,noise2,digit1,digit2,error}, denoted as 

\begin{equation}
	\label{gf}
	F(\mathbf{w}_{t})=\frac{1}{N}\sum_{i=1}^{N}F_{i}(\mathbf{w}_{t,i}),
\end{equation}
where $\mathbf{w}_{i}\in\mathbb{R}^{m\times 1}$ is the local parameters of the $i$-th device, and $\mathbf{w}_{t}=[\mathbf{w}_{t,1},\mathbf{w}_{t,2},\cdots,\mathbf{w}_{t,N}]\in\mathbb{R}^{m\times N}$ is the global parameter matrix. The local loss function $F_{i}(\cdot)$ and global loss function $F(\cdot)$ represent the loss function of the $i$-th device and the average loss function across all devices, respectively. However, if all devices only train their local models independently, the feasible domain of $\mathbf{w}_{i}$ only depends on $\mathcal{D}_{i}$. Therefore, for all local models, communication between devices can help to extend the feasible domain of $\mathbf{w}_{i}$ to $\mathcal{D}=\mathcal{D}_{1}\cup\mathcal{D}_{2}\cup\cdots\cup\mathcal{D}_{N}$, enabling each device to achieve the optimal value based on information from all devices. Therefore, we define the optimal local parameter $\mathbf{w}^{*}$ as

\begin{equation}
	\label{wstar}
	\mathbf{w}_{i}^{*} \triangleq \argmin\limits_{\mathbf{w}_{t,i}\in \mathbb{R}^{m}} F_{i}(\mathbf{w}_{t,i}),
\end{equation}
and $\mathbf{w}^{*} \triangleq [\mathbf{w}_{1}^{*},\mathbf{w}_{2}^{*},\cdots ,\mathbf{w}_{N}^{*}]\in\mathbb{R}^{m\times N}$.

\begin{figure}[!t]
	\vspace{-1cm}
	\centering
	\includegraphics[width=3.5in]{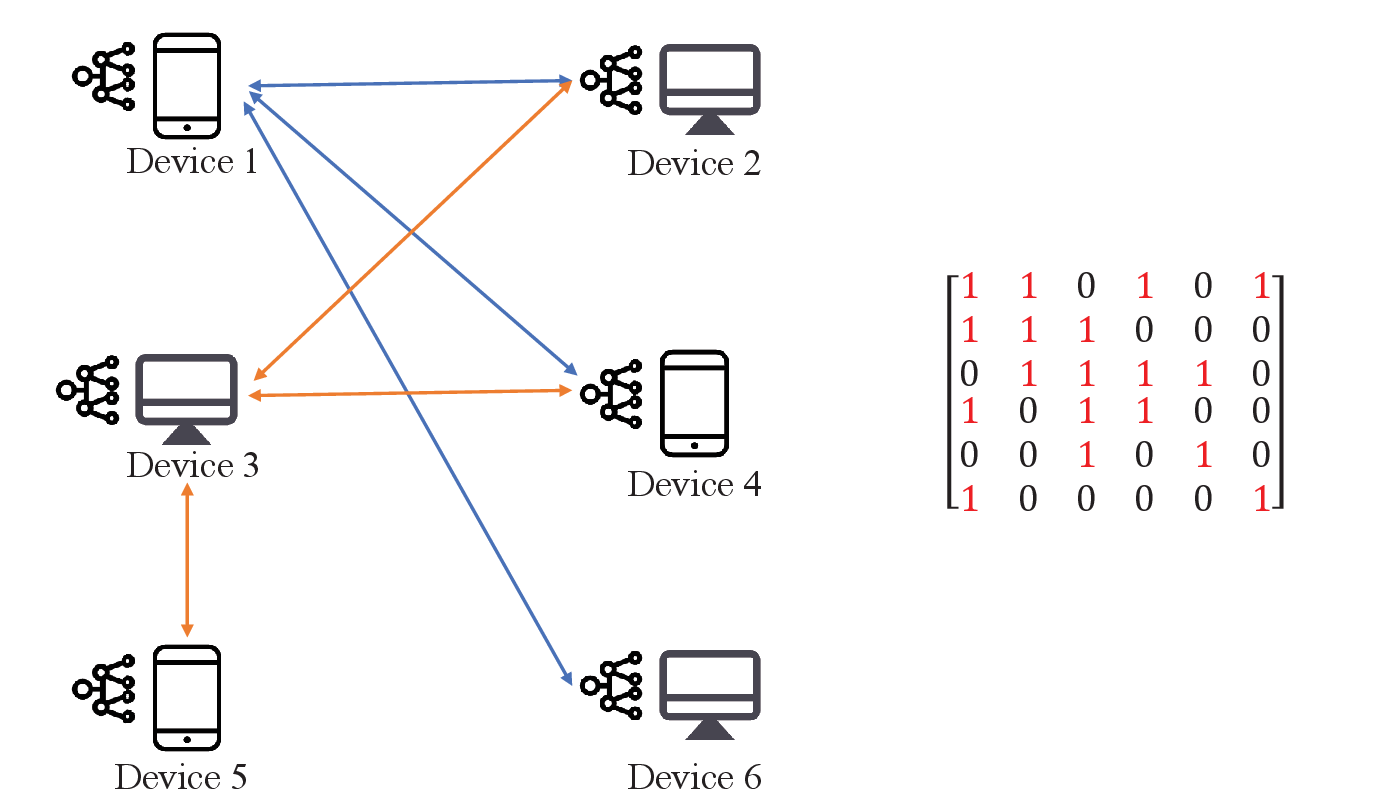}
	\label{fig1}
	\caption{An example of the graph.}
\end{figure}

Furthermore, in this paper, we consider the DFL framework with several rounds of local training and communication with each round of parameters update. Compared with previous works, which only include one round of local training or communication (e.g., \cite{con1,con2,con3,con4,noise1,noise2,digit1,digit2,error}), this framework is not only a more general case, but also more communication efficient \cite{dfl1,dfl2}. As shown in Fig. 2 and Algorithm 1, the process of parameters undergoing $\tau_{1}$ rounds of local training and $\tau_{2}$ rounds of communication is called an update round. Therefore, in the $(t+1)$-th update round, the initial parameter is $\mathbf{w}_{t}$ and the final parameter is $\mathbf{w}_{t+1}$. Specifically, when one device receives the parameters from all its neighbors, it starts one round of communication. There is a latency threshold on the whole communication period, which is large enough guarantee all devices can receive and aggregate the parameters after $\tau_{2}$ rounds of communication. Then, the next round of local training of all devices is allowed to get started. In addition, $\tilde{\mathbf{w}}_{t}$ denotes the parameters before the start of the communication process in the $(t+1)$-th update round. Besides, $\tilde{\mathbf{w}}_{t+\frac{i}{\tau_{1}}}$ denotes the results obtained in $i$-th training during the $\tau_{1}$ round of local training, and $\mathbf{w}_{t+\frac{j}{\tau_{2}}}$ denotes the results obtained in $j$-th communication during the $\tau_{2}$ round of communication, respectively. Based on this, we introduce local training model and communication model of DFL as shown in the following.

\begin{figure}[!t]
	\vspace{-1cm}
	\centering
	\includegraphics[width=3.5in]{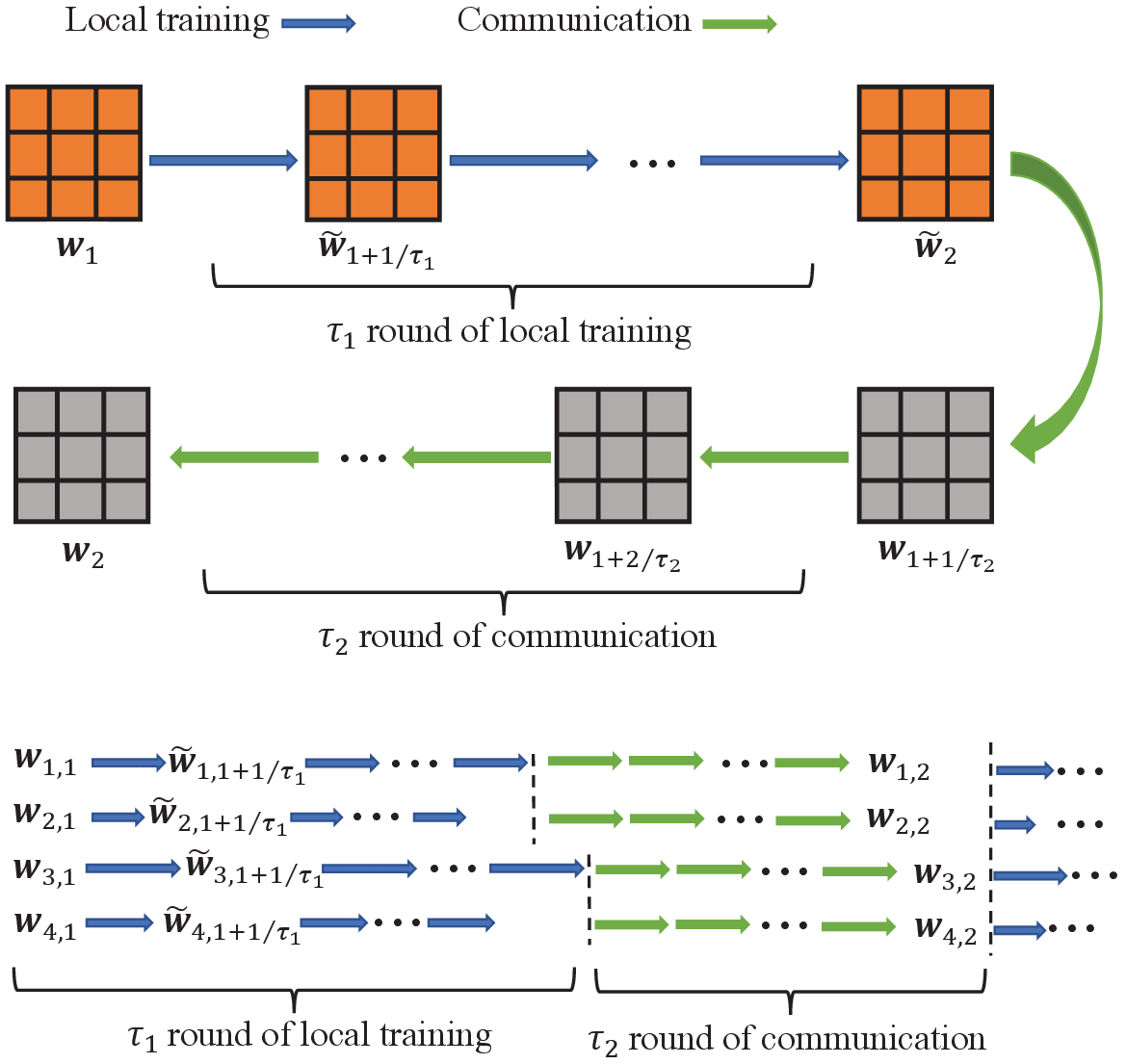}
	\label{fig2}
	\caption{Framework of multiple rounds of local training and communication}
\end{figure}

\begin{algorithm}
	\renewcommand{\algorithmicrequire}{\textbf{Input:}}
	\renewcommand{\algorithmicensure}{\textbf{Output:}}
	\caption{Decentralized federated learning (DFL)}
	\label{alg}
	\begin{algorithmic}[1]
		\REQUIRE $\tau_{1},\tau_{2},\mathbf{w}_{0}$
		\FOR{$t=1,2,\dots, T$}
		\FOR{$i=1,2,\dots, \tau_{1}$}
		\STATE For each device $k$ in parallel, its local parameter $\tilde{\mathbf{w}}_{t+\frac{i}{\tau_{1}},k}$ is updated by local training. 
		\ENDFOR
		\FOR{$j=1,2,\dots, \tau_{2}$}
		\STATE For each device $k$ in parallel, it communicates with its neighbors to exchange their parameters. 
		\ENDFOR
		\ENDFOR
		\ENSURE  $\mathbf{w}_{T}$
	\end{algorithmic}  
\end{algorithm}

\subsection{Local training model}

During the local training process, the parameters are usually updated by GD. We assume that in one update round, each user completes $\tau_{1}$ rounds of GD. Thus the parameter updated for local training can be expressed as
\begin{equation}
	\label{sgd}
	\tilde{\mathbf{w}}_{t+1} = \mathbf{w}_{t}-\sum_{i=1}^{\tau_{1}}\eta_{t}\nabla F\big(\tilde{\mathbf{w}}_{t+\frac{i}{\tau_{1}}}\big),
\end{equation}
where $\nabla F(\mathbf{w}_{t})=[\nabla F_{1}(\mathbf{w}_{t,1}),\cdots,\nabla F_{N}(\mathbf{w}_{t,N})]\in\mathbb{R}^{m\times N}$ is the global gradient matrix, and $F_{i}(\mathbf{w}_{t,i})\in\mathbb{R}^{m\times 1}$ is the local gradient of the $i$-th device. $\eta_{t}$ is the learning rate in $t$-th local training round.

\subsection{Communication model}

First, we consider the communication model without channel fading, noise and package error, which is an ideal communication model. In this case, we provide a simple framework to describe how the communication in DFL occurs. Based on this ideal case, we can directly obtain the communication model of digital and analog transmission by adjusting the specific form of parameters in the next sections. Besides, the ideal case can also provide a comparison benchmark of the specific transmission schemes. After $\tau_{1}$ rounds of local training, the DFL will conduct $\tau_{2}$ rounds of communication. Each device transmits its local model parameter to all neighbors via frequency-division multiple access (FDMA) with total bandwidth $B$, which is divided into $N$ sub-channels\footnote{FDMA is widely adopted in the FL and DFL frameworks (e.g., \cite{minE,minTE,minF}).}. Thus, the interference between devices are not considered in the system. In addition, different from the centralized BS with multiple antennas in FL, devices in DFL are usually mobile devices, such as mobile phones, computers or sensors, which are single-antenna devices \cite{minTE,sch1}. 
	
Moreover, in contrast to FL with a centralized BS, where the BS faces significant communication overhead due to a large number of devices connecting to it. In DFL, each device only connects to its own neighbors, which forms a smaller subset of devices in the system. Therefore, unlike FL with centralized BS, where advanced device selection algorithms might be necessary to select partial devices for communication efficiency, DFL typically does not require a specific device selection algorithm for each node. Thus, in each round of communication, the $i$-th device performs a weighted average of the received parameters by all connected devices and its own parameters, that is,
\begin{equation}
	\label{avg}
	\mathbf{w}_{t+\frac{j+1}{\tau_{2}},i}=\sum\limits_{k\in\mathcal{V}_{i}}\alpha_{k,i}\mathbf{w}_{t+\frac{j}{\tau_{2}},k},
\end{equation}
where $\mathcal{V}_{i}$ is the set of the devices connected to $i$-th device. $\alpha_{k,i}$ is the weight of aggregation of the $i$-th device, and $\sum_{k\in\mathcal{V}_{i}}\alpha_{k,i}=1$. The value of $\alpha_{i}$ is determined by the specific aggregation algorithm. Generally, the commonly used values are $\frac{1}{|\mathcal{V}_{i}|}$ or $\frac{|\mathcal{D}_{i}|}{\sum_{i\in\mathcal{V}_{i}}|\mathcal{D}_{i}|}$, where $|\mathcal{V}_{i}|$ is the number of neighbors of the $i$-th device and $|\mathcal{D}_{i}|$ is the size of its local dataset. Besides, there are some algorithms to improve the communication efficiency of the DFL system by designing the value of $\alpha_{i}$, such as \cite{con4}. After $\tau_{2}$ rounds of communication, the global parameter matrix of the DFL system can be summarized in the form of matrix multiplication, which is expressed as
\begin{equation}
	\label{boardcast}
	\mathbf{w}_{t+1}=\tilde{\mathbf{w}}_{t+1}\mathbf{P}^{\tau_{2}},
\end{equation}
where the $\mathbf{P}=[p_{ij}]_{N\times N}$ is mixing weights matrix. If there is a connection between the $i$-th device and the $j$-th one, $p_{ij}=\alpha_{j}$ and $p_{ji}=\alpha_{i}$, or $p_{ij}=p_{ji}=0$. In the next sections, we consider digital and analog transmission approaches applied in DFL and analyze their performance. The impact of package errors, channel fading and noise are also considered in the next sections.

\subsection{Preliminaries}

In order to complete the analysis of the performance of DFL with digital and analog transmission approaches, we make the following assumptions about the loss function and mixing weight matrix:

\subsubsection*{Assumption 1}

\textit{For each user and any pair of $\mathbf{w}$ and $\mathbf{w}'$, we assume the loss function is $\rho$-Lipscitz, i.e.,}
\begin{equation}
	\label{L}
	F(\mathbf{w}) - F(\mathbf{w}') \leq \rho \Vert \mathbf{w}-\mathbf{w}' \Vert_{\mathrm{F}},
\end{equation}
where $\Vert\cdot\Vert_{\mathrm{F}}$ is the Frobenius matrix norm. 

Assumption 1 is satisfied in many classic machine learning loss functions, such as cross entropy, logistic regression, etc. In addition, in the following assumption, we restrict the gradient of the loss function.
\subsubsection*{Assumption 2}

\textit{There is an upper bound of the Frobenius matrix norm of the gradient, i.e.,}
\begin{equation}
	\label{L}
	\Vert \nabla F(\mathbf{w}) \Vert_{\mathrm{F}}\leq G.
\end{equation}

This assumption is widely applied in the analysis of FL and DFL \cite{minE,minTE,minF}.
\subsubsection*{Assumption 3}

\textit{The mixing weights matrix $\mathbf{P}$ is a doubly stochastic matrix.}

This assumption is common in system design and analysis of DFL (e.g., \cite{noise1,noise2,error,dfl1,dfl2}). Based on the above assumptions, we can obtain some properties of $\mathbf{P}$, such as $\sum_{i=1}^{N}p_{ij}=\sum_{j=1}^{N}p_{ij}=1$, $\mathbf{1}^{T}\mathbf{P}=\mathbf{P1}=\mathbf{1}$ and $\mathbf{P}^{T}=\mathbf{P}$, where $\mathbf{1}\triangleq[1,1,\cdots,1]^{T}$. Thus, $\mathbf{P}$ is Markov matrices and their eigenvalue must be less than or equal to one.

\section{Performance Analysis on Digital Transmission}

In modern communication systems, digital transmission schemes are widely applied. There are several works about DFL consider the digital scheme to communicate with each other (e.g.,\cite{digit1,digit2,error}). In this section, we analyze the performance of DFL utilizing digital transmission over the wireless network, and investigate the impact of some system parameters and communication conditions on the proposed system. 

\subsection{Digital transmission mechanism}

In digital transmission, the parameters on devices are encoded as discrete digital signals. These signals are transmitted to their neighbors, and the receiving devices decode it to reconstruct the parameters. In the previous works about DFL with digital transmission approach, to ensure the reliability of communication between devices, TCP is widely used to transmit data packages over a communication network. It ensures the correctness of communication by resending erroneous packages. Thus, this protocol faces a huge communication overhead \cite{error}. However, limited communication resources are one of the most serious bottleneck of FL \cite{SURVEY,MMB,avg1,avg2}. Therefore, UDP, as a lightweight communication protocol with much less overhead, is considered in FL \cite{error,error2}. In the context of digital transmission, package errors refers to errors that occur during the transmission of parameters among each device. Each device's parameters are transmitted as a single package, and their data correctness can be verified using a mechanism known as the cyclic redundancy check (CRC) \cite{error2}. Package errors are caused by package loss or corruption during transmission and are considered random occurrences. In the case of UDP, unlike some heavyweight communication protocols like TCP, devices do not automatically resend erroneous packages \cite{error}. Since resending erroneous packages is unnecessary in UDP, communication failure caused by package errors need to be considered in DFL system. In the DFL system with package errors, whenever the received parameters of the $i$-th device contains package errors from its connected devices, the $i$-th device will not use these parameters to update its model \cite{error,error2}. Thus, \eqref{avg} can be written as
\begin{equation}
	\label{avg2}
	\mathbf{w}_{t+\frac{j+1}{\tau_{2}},i}=\sum\limits_{k\in\mathcal{V}_{i}}\alpha_{k,i}I(\mathbf{w}_{t+\frac{j}{\tau_{2}},k})\mathbf{w}_{t+\frac{j}{\tau_{2}},k},
\end{equation}
where $I(\mathbf{w}_{t+\frac{j}{\tau_{2}},k})=0$ indicates that the received parameters from the $k$-th device contains data errors and, hence, the $i$-th device will not use it to update its parameters, or $I(\mathbf{w}_{t+\frac{j}{\tau_{2}},k})=1$. As known, the occurrence of package errors is random. If we assume that the probability of correct transmission is $p$, we have $I(\mathbf{w}_{t+\frac{j}{\tau_{2}},k})\sim \mathcal{B}(1,p)$, and $\mathcal{B}(1,p)$ denotes the Bernoulli distribution with a parameter $p$, which is depended by the signal-noise-ratio (SNR) on the transmission link \cite{error2}. From \eqref{avg2}, we see that the probability of communication failure between 2 devices is $1-p$. Thus, the connectivity between devices can change over time due to the communication failure caused by package errors. Therefore, the mixing weights matrix with the erroneous communication can be denoted as a random matrix, and the graph of DFL system is a time-varying graph \cite{tv1,tv2,tv3}. Similar with \eqref{boardcast} and these works, we can also summary the updating of the DFL system with package errors in the form of matrix multiplication as follow:
\begin{equation}
	\label{boardcast11}
	\mathbf{w}_{t+1}=\tilde{\mathbf{w}}_{t+1}\prod_{k=1}^{\tau_{2}+k-1}\hat{\mathbf{P}}_{k},
\end{equation}
where $\hat{\mathbf{P}}_{k}\triangleq[\hat{p}_{ij}]$ and $\hat{p}_{ij}\triangleq p_{ij}I(\mathbf{w})$.

In summary, after all devices complete their local training periods, they broadcast packages of their parameters to the connected devices in several rounds. Since the impact of package errors, these parameters are randomly received by these devices.
\vspace{-3mm}
\subsection{Performance Analysis}

In the resource-constrained DFL with digital transmission approach, compared with TCP, to reduce communication overhead, devices represent their model parameters as digital messages for transmission with UDP. It is assumed that the receivers can decode these digital messages perfectly, and the error of the channel noise on the received parameters is negligible.\footnote{This assumption is widely applied in some works about FL (e.g., \cite{minE,error2}) and DFL (e.g., \cite{digit1,digit2,error}) with digital transmission. The effect of channel noise on performance will be discussed in the section on analog transmission. For the errors caused by imperfect decoding of digital signals, please refer to \cite{coding}.} Therefore, according to the communication process shown in \eqref{boardcast11}, we can use an upper bound to evaluate the performance of DFL with package errors caused by UDP with digital transmission. In addition, in the local training of the model, the learning rate is usually gradually reduced as the number of iterations increases. If we let $\eta_{t}\triangleq\frac{\eta}{(t-1)^{2}}$, where $\eta$ is the learning rate at the beginning of local training. To make the evaluation of the performance tractable and show the effects of the system parameters and communication conditions, we derive an upper bound of the gap between $F(\mathbf{w}_{T})$ and $F(\mathbf{w}^{*})$, where $F(\mathbf{w^{*}})$ means the minimum value of the loss function. The upper bound of resource-constrained DFL with digital communication can be given by the following theorem.

\textit{Theorem 1: The upper bound of the $\mathbb{E}[F(\bar{\mathbf{w}}_{T})-F(\mathbf{w}^{*})]$ is given by}
\begin{equation}
	\begin{aligned}
		\label{gap1}
		\mathbb{E}[F(\mathbf{w}_{T})&-F(\mathbf{w}^{*})]\leq\underbrace{\rho\Vert\mathbf{w}_{1}\Vert_{\mathrm{F}}\sqrt{(\Vert\mathbf{Q}_{T-1}\Vert_{\mathrm{F}}^{2}-N)p^{2}+Np}}_{A_{1}(p): \text{Impact of communication}}\\
		&+\underbrace{\rho\eta\tau_{1}A_{1}(p)\Big(\frac{1}{\sqrt{N(T-1)}}+\frac{N}{T-1}\Big)}_{A_{2}(p,T): \text{Impact of local training}}+\rho\Vert\mathbf{w}^{*}\Vert_{\mathrm{F}},
	\end{aligned}
\end{equation}
where $\mathbf{Q}_{j}\triangleq\mathbf{P}^{j\tau_{2}}$ and $p$ is the probability of correct communication of one communication round.

\textit{Proof: See Appendix A.$\hfill\qedsymbol$}

\textit{Remark 1: Since $\mathbf{P}$ is a Markov matrix, so is $\mathbf{Q}_{T-1}=[q_{ij,T}]_{N\times N}$. Thus, based on Markov Chain Stationary State Theorem, $\lim_{T \to +\infty}\mathbf{Q}_{T-1}=\frac{1}{N}\mathbf{1}\mathbf{1}^{T}$. Besides, since $\sum_{i=1}^{N}\sum_{j=1}^{N}q_{ij,T}=N$, we have $\Vert\mathbf{Q}_{T}\Vert_{\mathrm{F}}^{2}=\sum_{i=1}^{N}\sum_{j=1}^{N}q_{ij,T}^{2}\geq \Vert\frac{1}{N}\mathbf{1}\mathbf{1}^{T}\Vert_{\mathrm{F}}^{2}$. Thus, a larger $\tau_{2}$ means a smaller value of $\Vert\mathbf{Q}_{T}\Vert_{\mathrm{F}}^{2}$.}

Note that, Theorem 1 provides a convergence bound of DFL under the digital transmission with UDP. Although the convergence of DFL has been discussed in some previous works \cite{digit1,digit2,error,dfl1}, our results still provide some different insights. According to Theorem 1, we can divide the factors that affect the performance of DFL with erroneous communication with digital transmission approach into two categories: system parameters and communication conditions. 

On one hand, the system parameters include number of local training rounds and communication rounds (e.g., $\tau_{1}$ and $\tau_{2}$), number of devices in DFL (e.g., $N$) and the number of update rounds that DFL can complete (e.g., $T$). First, we assume that the training resources and communication resources are independent of each other. Based on this bound, a lower $\tau_{1}$ can make DFL perform better. This conclusion is consistent with the previous works regarding the DFL \cite{dfl1} and distributed optimization \cite{dopt}. The reason is that what DFL needs to find is the global optimum based on all device data sets, which requires more communication rounds between devices to exchange information. In contrast, local updates via gradients computed on the device's own data do not utilize data from other devices. This is due to the fact that DFL may only obtain local optima for some devices based on their own data sets. On the other hand, since the value range of $p$ is $(0.5,1]$, both $A_{1}$ and $A_{2}$ are decreasing functions with respect to $p$ in this range.
Thus, the DFL system can achieve a better performance with a lower probability of communication failure caused by package errors. 

In the mentioned previous works \cite{digit1,digit2,error,dfl1}, since the recourse constraint are not considered in \cite{digit1,digit2}, to achieve a better performance, the case they studied is $\tau_{1}=1$ and transmission with TCP. Thus, their results cannot reveal the impact of $\tau_{1}$ and $p$ in the cost constrained case. In addition, even though the cases with UDP and multiple local training rounds are discussed in \cite{error} and \cite{dfl1} respectively. However, based on Theorem 1, $p$ and $\tau_{1}$ are coupled in the convergence bound. Therefore, jointly considering these two cases in one scenario gives a more general and practical insight of DFL.
\vspace{-3mm}
\subsection{Allocation of resources between computation and communication in DFL}

In this section, considering the computation and communication costs are not independent, we analyze how to allocate the costs between computation and communication in DFL to enhance the model performance. In the previous work of DFL with recourse constraints \cite{dfl1}, independent computation and communication costs is implicitly assumed. Based on this assumption, $\tau_{1}=1$ is the best choice to achieve a better model performance, which is also be confirmed by our discussion after Theorem 1. Thus, some works set $\tau_{1}=1$ in their system to obtain a higher accuracy directly (e.g., \cite{dopt,digit1,digit2}). However, coupled computation and communication costs is a more practical case. In this case, $\tau_{1}=1$ may not be the optimal setting. For example, the energy of one device need to allocate to several rounds of local training and communication in one update round to finish the model training. If we set $\tau_{1}=1$, it will lead to a higher frequency of \textit{$\tau_{2}$ rounds of communication}. Thus, the total energy budget may be consumed faster, which in turn leads to a smaller $T$, making the final performance worse. Therefore, obtaining a balanced $\tau_{1}$ and $\tau_{2}$ by allocating of resources between computation and communication to achieve a better model performance is a critical issue in DFL.

Since the value of the upper bound in \eqref{gap1}, which decreases monotonically as $T$ increases, is used to capture the model performance. We can formulate a optimization problem to achieve the goal of resources allocation between computation and communication in DFL.
\begin{align}
	&\min_{\tau_{1},\tau_{2}}\quad A_{1}(p)+A_{2}(p,T) \label{P2}\\
	&\;\textrm{s.t.}\quad \tau_{1}R_{1}+\tau_{2}R_{2}= R_{c}, \tag{\ref{P2}{a}} \label{P1a}
\end{align}
where the total resources of DFL is $R_{c}$, and the costs of local training and communication are $R_{1}$ and $R_{2}$ respectively.

\textit{Remark 4: $R_{c}$ can represent different resources, such as time, energy and so on. For example, if the $R_{1}$ and $R_{2}$ denote the computation and communication time or energy respectively, $R_{1}$ depends on CPU frequency and CPU cycle, and $R_{2}$ depends on transmission power and channel condition. The specific calculation formula can refer to \cite{minE,minTE,minF,error2}.}

\textit{Corollary 1: The optimal $\tau_{1}$ and $\tau_{2}$ in \eqref{P2} is given by}
\begin{equation}
	\begin{aligned}
		\label{optt}
		\begin{cases}
			\tau_{2}^{*}=\lfloor \frac{R_{c}}{R_{2}} \rfloor\\
			\tau_{1}^{*}=\max\big\{1,\lfloor\frac{R_{c}-\tau_{2}^{*}R_{2}}{R_{1}}\rfloor\big\}.
		\end{cases}
	\end{aligned}
\end{equation}

\textit{Proof: This result is obtained by linear programming.$\hfill\qedsymbol$}

From this corollary, we know that, the optimal allocation scheme between computation and communication is setting $\tau_{2}$ as large as possible, and the remaining resources is for local training. More communication rounds can synchronize the local parameters on each device to more devices faster to achieve the performance improvement. Thus, in the DFL shown in Algorithm 1, we can estimate $R_{1}$ and $R_{2}$ based on the computing capacity of the device and the communication conditions of the system and set the budget of recourse cost as $R_{c}$. Then calculate $\tau_{1}^{*}$ and $\tau_{2}^{*}$ as the input of Algorithm 1, to apply this allocation scheme on DFL.
\subsection{Convergence Rate Analysis}

By calculating the minimum number of update rounds required for DFL to reach the convergence condition, its convergence rate can be measured. Since the time-varying matrix is a random variable, we consider the following condition as the convergence condition of DFL.

\textit{Definition 1 (Convergence condition): After $T$ rounds of updating, if for each device, its $\bar{\mathbf{w}}_{T}$ satisfies the following condition,}
\begin{equation}
	\label{con}
	\mathrm{Pr}\Big(\frac{\Vert\mathbf{w}_{i,T}-\mathbf{w}_{i}^{*}\Vert_{\mathrm{2}}}{\Vert\mathbf{w}_{i,1}\Vert_{\mathrm{2}}+\delta}\geq\epsilon\Big)\leq \epsilon,\quad\forall i,
\end{equation}
\textit{it is converged.}

\textit{Remark 2: Note that $\Vert\bar{\mathbf{w}}_{i,1}\Vert_{\mathrm{2}}$ may be equal to $0$, $\delta$ is a constant used to ensure that the denominator is not $0$ and $\frac{\Vert\mathbf{w}_{i,T}-\mathbf{w}_{i}^{*}\Vert_{\mathrm{2}}}{\Vert\mathbf{w}_{i,1}\Vert_{\mathrm{2}}+\delta}<1$.}

This condition aims to ensure that the difference between each local parameters and the global optima remains bounded by a threshold $\epsilon$, which is normalized by $\Vert\bar{\mathbf{w}}_{i,1}\Vert_{\mathrm{2}}$. By maintaining this bounded difference with a high probability for all devices, the convergence condition aims to guarantee the convergence of the local models towards the global optimal model. Moreover, due to the inherent randomness of package error in the DFL with digital transmission, this convergence condition is expressed by the probability notation. This condition is also adopted in \cite{gossip}.

\textit{Definition 2 (Density of connections): For a fixed graph topology, the second largest eigenvalue (e.g., $\lambda_{2}$) of the corresponding graph matrix can evaluate its connectivity density \cite{gossip}. A denser graph topology has a lower $\lambda_{2}$. Thus, we design a new variable, beta, to measure the average connection density of time-varying graph topology.}
\begin{equation}
	\label{lambda}
	\bar{\beta}=\sqrt{\frac{\sum_{i=1}^{T-1}\beta_{i}}{T-1}},
\end{equation}
where $\beta_{i}=\frac{1}{1-\lambda_{2}(\hat{\mathbf{Q}}_{i}^{T}\hat{\mathbf{Q}}_{i})}$ and $\hat{\mathbf{Q}}_{i}\triangleq\prod_{j=1}^{i\tau_{2}}\hat{\mathbf{P}}_{i}$.

\textit{Remark 3: Since $\hat{\mathbf{Q}}_{i}^{T}\hat{\mathbf{Q}}_{i}$ is also a Markov matrix, its maximal eigenvalue is always 1 and $\lambda_{2}(\hat{\mathbf{Q}}_{i}^{T}\hat{\mathbf{Q}}_{i})<1$ reflects the density of connections between devices. Thus, $\beta_{i}$ can also evaluate this density since it has the same variation trend as $\lambda_{2}(\hat{\mathbf{Q}}_{i}^{T}\hat{\mathbf{Q}}_{i})$. Based on this, the average density of time-varying graph topology can be measured by $\bar{\beta}$.}

According to the convergence condition, the following result characterizes the convergence rate of DFL in time-varying graph topology.

\textit{Theorem 2: When $\eta_{t}=\frac{\eta}{(T-1)^{\frac{3}{2}}}$, the number of update rounds $T$ required to reach the convergence condition is bounded by}
\begin{equation}
	\begin{aligned}
		\label{rate}
		T\geq\frac{\phi_{1}(\tau_{1},\bar{\beta})}{\epsilon^{2}-\phi_{2}(p,N,H)}+1,
	\end{aligned}
\end{equation}
where
\begin{equation*}
	\begin{cases}
		\phi_{1}(\tau_{1},\bar{\beta})\triangleq\delta^{-1}\eta\tau_{1}G\bar{\beta},\\
		\phi_{2}(p,N,H)\triangleq\sqrt{(\Vert\mathbf{Q}_{T-1}\Vert_{\mathrm{F}}^{2}-N)p^{2}+Np+N}+H.
	\end{cases}
\end{equation*}
and $H\triangleq\frac{\Vert\bar{\mathbf{w}}_{i,1}-\mathbf{w}_{i}^{*}\Vert_{\mathrm{2}}}{\Vert\bar{\mathbf{w}}_{i,1}\Vert_{\mathrm{2}}}$ reflects the influence of the initial value of the parameter, which is usually random.

\textit{Proof: See Appendix B.$\hfill\qedsymbol$}

Theorem 2 reflects that, the number of devices, probability of correct communication, selection of initial values, and graph topology jointly determine the convergence rate of DFL with package errors. Specifically, $\phi_{1}(\tau_{1},\bar{\beta})$ implies the negative effects of adding more local training rounds and making the connections between devices more sparse, since $\phi_{1}$ is an increasing function of $\tau_{1}$ and $\bar{\beta}$. From the local training perspective, this result shows that employing more local training and fewer communication causes worse performance as a result of a growing discrepancy among local models due to less synchronization frequency. From the graph topology perspective, based on Definition 2, we note that, when the aggregation algorithms are the same, a larger value of $\bar{\beta}$ means that the graph topology composed of devices is sparser. Thus, a sparser graph slows down the convergence since $\phi_{1}$ is an increasing function of $\bar{\beta}$. This also means that, a more densely connected graph topology can achieve a better model performance under the same $T$. On the contrary, by observing $\phi_{2}(p,N,H)$, we can see that, more number of devices (e.g., a larger $N$) and a higher probability of correct communication (e.g., a larger $p$) speed up the convergence of DFL. Besides, the closer the initial value of the parameter to the optimal value (e.g., a smaller $H$), the faster the convergence rate. 

Next, by observing the form of Theorem 2, we note that, DFL converges in this case requires $\epsilon^{2}-\phi_{2}(p,N,H)$ is positive. It means that, in DFL with package errors, if the probability of communication failure is high enough which causes $p$ too low to allow $\epsilon^{2}-\phi_{2}(p,N,H)>0$, it cannot converge in limited update rounds with resource constraints. Thus, we also investigate and study the minimum probability of correction communication that can be accepted in DFL. The result is obtained by the following corollary.

\textit{Corollary 2: The minimum probability of correction communication that DFL can tolerate is given by}
\begin{equation}
	\label{p}
	p\geq\frac{-N+\sqrt{N^{2}+4(\Vert\mathbf{Q}_{T-1}\Vert_{\mathrm{F}}^{2}-N)(\epsilon^{2}-H)^{2}}}{2(N-\Vert\mathbf{Q}_{T-1}\Vert_{\mathrm{F}}^{2})}.
\end{equation}

\textit{Proof: Let $\epsilon^{2}-\phi_{2}(p,N,H)>0$ and we complete it. $\hfill\qedsymbol$}

From Corollary 2, we know that, with the number of devices increasing in DFL, the lower the correct probability of communication is to ensure the convergence of DFL. Because in the same graph topology, more devices in DFL causes more links between devices. Therefore, the possibility of errors caused by package errors in all communications at the same time is reduced.

In addition, the communication complexity can be analyzed based on Theorem 1. Communication complexity can be defined as defined as the number of $T$ needed to achieve the convergence condition \cite{con1}. In our DFL system, it is given by the following corollary.

\textit{Corollary 3: When $\delta=\epsilon^{-1}$, the communication complexity is obtained as follows:}
\begin{equation}
	\begin{aligned}
		\label{plex}
		T=\mathcal{O}\Big(\frac{\epsilon}{\epsilon^{2}-\phi_{2}(p,N,H)}\Big).
	\end{aligned}
\end{equation}
Similar to Theorem 2, the influence of $p$ and $N$ on the communication complexity is coupled. More devices and a higher error probability can increase the communication complexity of the system. Besides, the impact of $p$ on the communication complexity is more obvious in the DFL system with larger $N$. Furthermore, note that, when $p=1$, $N$ and $H$ are small, communication complexity is approximately $\mathcal{O}(\frac{1}{\epsilon})$. This means that linear convergence is a lower bound on the communication complexity of DFL, which is consistent with the analysis of previous works on DFL, such as \cite{error,dfl1}.

\section{Performance Analysis on Analog Transmission}

Although digital communication is widely used in various scenarios, some works still focus on the analog transmission over FL \cite{analog1,analog2,analog3,analog4} and DFL frameworks \cite{noise1,noise2}, such as over-the-air computation \cite{analog2,analog3}, low-resolution digital-to-analog converters \cite{analog4} and DFL with differential privacy \cite{noise1,noise2}. Since in a cost-constraint wireless network, compared with digital scheme with complex encoding, decoding, and signal processing algorithms, analog communication techniques typically involve simpler modulation and demodulation schemes to reduce the computation costs \cite{analog2}. Besides, some analog schemes are more bandwidth-efficient since certain computations can be carried out naturally through channels, resulting from waveform-superposition property of wireless signals \cite{analog3}, and complex signaling schemes used for encoding and error correction are omitted in analog communication, which makes more efficient use of the limited available spectrum and mitigates potential bandwidth constraints \cite{analog1,analog3}. Moreover, analyzing the performance of DFL on analog transmission can provide a comparison benchmark of the digital case. This comparison is finished in the subsection D.

\subsection{Analog transmission mechanism}

With analog transmission, devices directly transmit their respective parameters, which does not employ any digital code, either for compression or channel coding. Thus, in the analog transmission protocol, communication between devices does not suffer from package errors, but the received parameters are affected by the channel fading and noise. Specifically, the block fading channel model is adopted in this paper, i.e., channels remain unchanged during each communication round, and are different across different communication rounds following by the Rayleigh distribution.

In DFL with analog transmission approach, to utilize both the real and imaginary components of the available subchannel, the $i$-th device transforms the real parameter vector $\mathbf{w}_{t+\frac{j}{\tau_{2}},i}$ to a complex vector $\mathbf{x}_{i,j}$ and broadcast it to the connected devices in an analog manner. In particular, one of the most popular transforming schemes is to treat the first half of $\mathbf{w}_{t+\frac{j}{\tau_{2}},i}$ as the real part of $\mathbf{x}_{i,j}$ and the rest as the imaginary part. Thus, $\mathbf{x}_{i,j}$ can be expressed as
\begin{equation}
	\begin{aligned}
		\label{complex}
		\mathbf{x}_{i} =  [w_{i,1}+jw_{i,\frac{m}{2}+1},\cdots,w_{i,\frac{m}{2}}+jw_{i,m}]^{T}\in\mathbb{C}^{\frac{m}{2}},
	\end{aligned}
\end{equation}
where $w_{i,1},\cdots,w_{i,m}$ are the values of $\mathbf{w}_{t+\frac{j}{\tau_{2}},i}$ and $j=\sqrt{-1}$ is the imaginary unit. Therefore, The received signal at the connected devices is given by
\begin{equation}
	\begin{aligned}
		\label{receive}
		\mathbf{y}_{i,j} = \sqrt{P_{i}d_{i,j}^{-\gamma}}h_{i,j}\mathbf{x}_{i,j}+\mathbf{z}_{i,j},
	\end{aligned}
\end{equation}
where $P_{i}$ is the transmission power of the $i$-th device, the fading coefficient $h_{i,j}\in\mathbb{C}$ remains unchanged within one communication block. $d_{i,j}$ denotes the distance between two connected devices and $\gamma$ is the path loss exponent factor. $\mathbf{z}_{i,j}=[z_{1,ij},\cdots,z_{\frac{m}{2},ij}]^{T}\in\mathbb{C}^{\frac{m}{2}}$ is the channel noise between the $i$-th device and its neighbors in the $j$-th communication round, and $z_{k,ij}\sim\mathcal{CN}(0,\sigma_{z}^{2})$, $k=1,2,\cdots, \frac{m}{2}$. After the signal $\mathbf{y}_{i,j}$ received by the devices, it will be recovered by the channel inversion method, which is widely adopted in FL with analog transmission (e.g., \cite{fading1,fading2}). Thus, based on \eqref{complex} and \eqref{receive} the received parameter vector can be expressed as
\begin{equation}
	\begin{aligned}
		\label{receive2}
		\hat{\mathbf{w}}_{t+\frac{j}{\tau_{2}},i} = [Re(\hat{\mathbf{y}}_{i,j}^{T}),Im(\hat{\mathbf{y}}_{i,j}^{T})]^{T}= \mathbf{w}_{t+\frac{j}{\tau_{2}},i}+\tilde{\mathbf{n}}_{i,j},
	\end{aligned}
\end{equation}
where
\begin{equation}
	\begin{aligned}
		\label{receive3}
		\hat{\mathbf{y}}_{i,j}=\kappa \mathbf{y}_{i,j}=\frac{P_{i}^{-\frac{1}{2}}d_{i,j}^{\frac{\gamma}{2}}h_{i,j}^{H}}{|h_{i,j}|^{2}}\mathbf{y}_{i,j},
	\end{aligned}
\end{equation}
$\kappa$ is the post-processing factor and $\tilde{\mathbf{n}}_{i,j}=\kappa\mathbf{n}_{i,j}$, and $\mathbf{n}_{i,j}=[n_{1,ij},\cdots,n_{m,ij}]^{T}=[Re(\mathbf{z}_{i,j}^{T}),Im(\mathbf{z}_{i,j}^{T})]^{T}$, $Re(\cdot)$ and $Im(\cdot)$ are real vectors composed of the real and imaginary parts of the complex vector, respectively. Thus, $n_{k,i}\sim\mathcal{N}(0,\sigma_{n}^{2})$ and $\sigma_{n}^{2}=\sigma_{z}^{2}/2$, where $k=1,2,\cdots, m$. Thus, $\tilde{n}_{k,i}\sim\mathcal{N}(0,\kappa^{2}\sigma_{n}^{2})$, which is the element of $\tilde{\mathbf{n}}_{i,j}$.

Since the channel coding is not applied in the analog communication, package errors do not exist in this transmission approach. Thus, different from the communication model of digital transmission in \eqref{boardcast11}, the mixing weights matrix in analog transmission is modeled as a fixed matrix. Furthermore, according to \eqref{receive2}, the global parameter matrix of the DFL system with analog transmission protocol, which is updated after $\tau_{2}$ rounds of communication, can be written as
\begin{equation}
	\begin{aligned}
		\label{boardcast2}
		\hat{\mathbf{w}}_{t+1}&=\big(\mathbf{w}_{t+\frac{\tau_{2}-1}{\tau_{2}}}+\tilde{\mathbf{n}}_{\tau_{2}}\big)\mathbf{P}\\
		&=\mathbf{w}_{t+\frac{\tau_{2}-1}{\tau_{2}}}\mathbf{P}+\tilde{\mathbf{n}}_{\tau_{2}}\mathbf{P}\\
		&=\big(\mathbf{w}_{t+\frac{\tau_{2}-2}{\tau_{2}}}+\tilde{\mathbf{n}}_{\tau_{2}-1}\big)\mathbf{P}^{2}+\mathbf{n}_{\tau_{2}}\mathbf{P}\\
		&=\tilde{\mathbf{w}}_{t+1}\mathbf{P}^{\tau_{2}}+\sum_{j=1}^{\tau_{2}}\tilde{\mathbf{n}}_{j}\mathbf{P}^{\tau_{2}+j-1},
	\end{aligned}
\end{equation} 
where $\tilde{\mathbf{n}}_{j}=[\tilde{\mathbf{n}}_{1,j},\cdots,\tilde{\mathbf{n}}_{N,j}]=[\tilde{n}_{kl,j}]_{m\times N}$ denotes the noise matrix in the $j$-th communication, $l=1,2,\cdots,N$ is the index of device.
\subsection{Performance Analysis}

In analog transmission approach, since the devices broadcast their parameters to the connected devices in an uncoded manner, after $\tau_{2}$ rounds of communication, the error on the receivers caused by noise accumulation is revealed in \eqref{boardcast2}. Therefore, similarly to the Theorem 1, the upper bound of the $\mathbb{E}[F(\hat{\mathbf{w}}_{T})-F(\mathbf{w}^{*})]$ to evaluate the performance of DFL with analog communication can be given by the following theorem.
\textit{Theorem 3: The upper bound of the $\mathbb{E}[F(\hat{\mathbf{w}}_{T})-F(\mathbf{w}^{*})]$ is given by}
\begin{equation}
	\begin{aligned}
		\label{gap4}
		\mathbb{E}[F(\hat{\mathbf{w}}_{T})-F(\mathbf{w}^{*})]&\leq A_{1}(1)+A_{2}(1,T)\\
		&+\underbrace{(T-1)\rho\kappa\sigma_{n}\sqrt{\tau_{2}mN}}_{A_{3}(\kappa,\sigma_{n}): \text{Impact of the fading channel}}
	+\rho\Vert\mathbf{w}^{*}\Vert_{\mathrm{F}},
	\end{aligned}
\end{equation}
where $A_{1}(p)$ and $A_{2}(p,T)$ are defined in \eqref{gap1}.

\textit{Proof: See Appendix C. $\hfill\qedsymbol$}

From this theorem, we know that, more update rounds or communication rounds (e.g., a larger $T$ or $\tau_{2}$) makes a lower $A_{2}$ and $A_{1}$, which lead to a better performance, but with more communication rounds and error accumulation caused by channel fading and noise. Moreover, we remark $A_{1}$ and $A_{2}$ decrease with $\tau_{2}$, while $A_{3}$ is an increasing function of $\tau_{2}$. Thus, the impact of $\tau_{2}$ on the performance improvement in this general case is not evident.However, for a more serious error accumulation caused by channel fading and noise, which means larger $\kappa$, $\sigma_{n}$ and $T$, the destructive effect of increasing $\tau_{2}$ on $A_{3}$ becomes more significant. However, the reduction in $A_{1}$ and $A_{2}$ are the same as having less error caused by fading and noise, where we note that they are not the functions of $\kappa$ and $\sigma_{n}$. Therefore, for a more serious error accumulation, where $A_{3}$ is more sensitive to the change in $\tau_{2}$, designing an efficient $\tau_{2}$ is more necessary. This corroborates our intuitive understanding of performance improvement in DFL with fading channels, where for a more serious error accumulation, increasing the number of communication round excessively leads to more errors and makes a worse performance. Thus, $\tau_{2}$ must be balanced to improve the performance. Besides, more devices in DFL (e.g., a larger $N$) means more links between devices. So in a round of communication, there are more devices communicating with each other, which also causes more serious error accumulation (e.g., a higher $A_{3}$). However, training with more data from more devices can also improve the model performance (e.g., a smaller $A_{1}$).
\subsection{Impact of fading channels}

Note that the impact of channel fading, noise and the impact of communication/local training are independent of each other in \eqref{gap1}. In this subsection, we further discuss the influence of fading channels on the DFL. From the perspective of performance, we study the impact of fading channel and noise on the model performance by deriving the gap between the loss function of DFL after $T$ rounds of updates over the fading channel and the loss function in the case of perfect communication.

\textit{Theorem 4: The upper bound of $\mathbb{E}[F(\hat{\mathbf{w}}_{T})-F(\mathbf{w}_{T})]$ is}
\begin{equation}
	\label{gap2}
	\mathbb{E}[F(\hat{\mathbf{w}}_{T})-F(\mathbf{w}_{T})]\leq
	(T-1)\rho\kappa\sigma_{n}\sqrt{\tau_{2}mN},
\end{equation}

\textit{Proof: The proof follows from (45) in Appendix C, and thus omitted for brevity.$\hfill\qedsymbol$}

According to this bound, we know that, more update rounds, communication rounds and devices all lead to the accumulation of error caused by channel fading and noise on the DFL. However, these 3 factors (e.g., $T$, $\tau_{2}$ and $N$) of error accumulation are different. More devices in DFL means that, channel fading and noise cause more serious errors in a round of communication, because there are more links between devices. In addition, a larger $\tau_{2}$ reflects more communication rounds in one update round, which accumulates more errors. Besides, larger $T$ causes an increase in total communication rounds by more update rounds, so DFL produces more significant errors due to more severe error accumulation caused by channel fading and noise.

In addition, too much error caused by channel fading and noise can cause DFL to fail to converge. The maximum error in one communication round that DFL can tolerate is given by the following corollary.

\textit{Corollary 4: The maximum error in one communication round} that DFL can tolerate is given by
\begin{equation}
	\label{noise}
	\kappa\sigma_{n}\leq\frac{\eta\tau_{1}A_{1}(1)}{\rho\sqrt{\tau_{2}m}}\Bigg(\frac{(T-1)^{1.5}}{2N}+\frac{
	\sqrt{N}}{(T-1)^{2}}\Bigg).
\end{equation}

\textit{Proof: The proof follows by solving $\frac{\partial (A_{1}+A_{2}+A_{3})}{\partial T}\leq 0$, and thus omitted for brevity. $\hfill\qedsymbol$}

Note that, while the system has a fixed $T$ and $N$, this bound only depends on $\tau_{1}$ and $\tau_{2}$, where smaller $\tau_{1}$ or larger $\tau_{2}$ makes this bound lower. Because as $\tau_{2}$ increases, more errors are accumulated. In addition, since the limited resources of computation and communication in DFL, with $\tau_{1}$ increasing, less resources are utilized for devices to communicate, which results in less error accumulation.
\subsection{Performance Comparison}

In this subsection, we are going to compare the performance of three different communication scenarios. Specifically, the scenarios compared include:

\begin{itemize}
	
	\item{DFL with analog communication protocol}
	
	\item{DFL with digital communication protocol and UDP}
	
	\item{DFL with digital communication protocol and TCP} 
	
\end{itemize}

The comparison results are summarized in the following corollary.

\textit{Corollary 5: For the above three scenarios, the relationship of the upper bound is given by}
\begin{equation}
	\label{comp}
	\begin{aligned}
		&A_{1}(1){+}A_{2}(1,T){+}A_{3}(\kappa,\sigma_{n}){>}A_{1}(1){+}A_{2}(1,T){+}A_{3}(P_{i}^{\frac{1}{2}},0)\\
		&A_{1}(p){+}A_{2}(p,T){+}A_{3}(P_{i}^{\frac{1}{2}},0){>}A_{1}(1){+}A_{2}(1,T){+}A_{3}(P_{i}^{\frac{1}{2}},0).
	\end{aligned}
\end{equation}

\textit{Proof: See Appendix D. $\hfill\qedsymbol$}

From this Corollary, we know that, compared with using analog transmission or digital transmission in resource-constrained DFL, if the communication resources of each device in the DFL system are sufficient, the use of digital transmission with TCP can achieve a better model performance.

\section{Simulation Results}
\subsection{System Setup}
\textit{1) DFL system:} For our simulations, we consider a DFL system with digital and analog transmission approaches. Besides, we consider three different connection topologies between devices, namely chain, ring and full connection. In particular, we study the cases with 4, 8 and 12 devices in the DFL system respectively. Different noise power in analog communication,  probability of correction communication in digital communication and learning rates are also considered in our simulations. In the case of digital communication, we focus on the effect of system parameters and probability of correct communication on performance. In the case of analog communication, we discuss the effect of noise power on model convergence with different system parameters.

\textit{2) Models:} In our simulations, each user independently trains a CNN model with the same structure, and completes the task of image classification on different data sets.

\textit{3) Datasets:} Two datasets commonly used in image classification, fashion-MNIST and CIFAR-10 are applied in our simulations. Both of them have 10 classes of images. Besides, for both datasets, we split them into training and testing sets consisting of 50,000 and 10,000 images, respectively.

\subsection{Results and Discussions}
In this subsection, we validate the above analysis by comparing the performance and convergence rate of CNN models trained in the DFL system with different parameters and different scenarios.

First, we consider the case of DFL without channel noise. In this case, we explore the impact of $\tau_{1}$ and $\tau_{2}$ with/without resource constraints on the model performance, and the impact of $N$ and $\lambda_{2}$ on the convergence rate.

\textit{1) DFL \textbf{without} resource constraints in digital transmission:} In this case, we compare the model performance with different number of local training rounds (e.g., $\tau_{1}$) and communication rounds (e.g., $\tau_{2}$) in erroneous communication. The results are shown in Fig. 3(a) and Fig. 3(b).

\begin{figure*}[!t]
	\vspace{-1.5cm}
	\centering
	\subfloat[]{\includegraphics[width=2.4in]{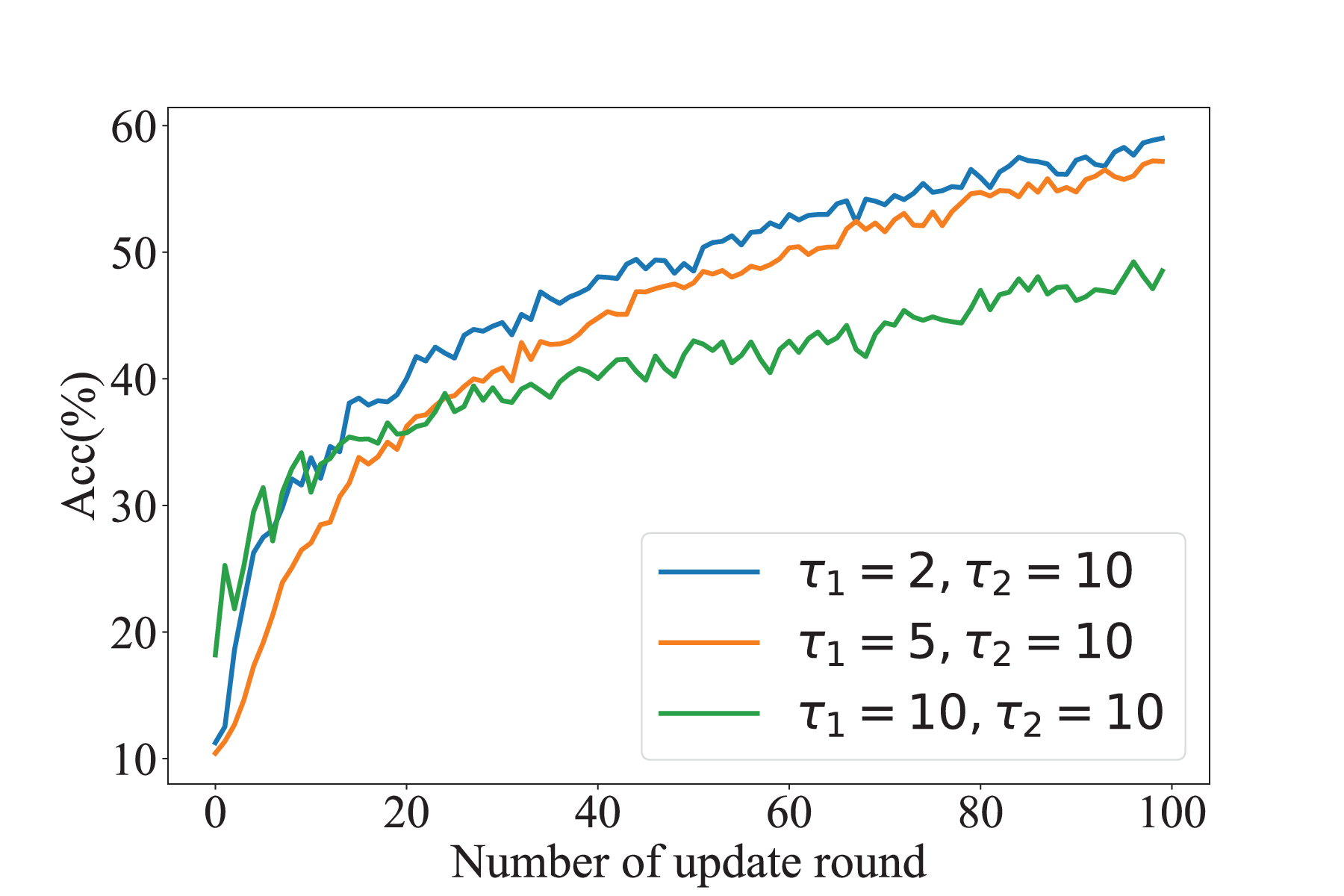}
		\label{fig3_1}}
	\subfloat[]{\includegraphics[width=2.4in]{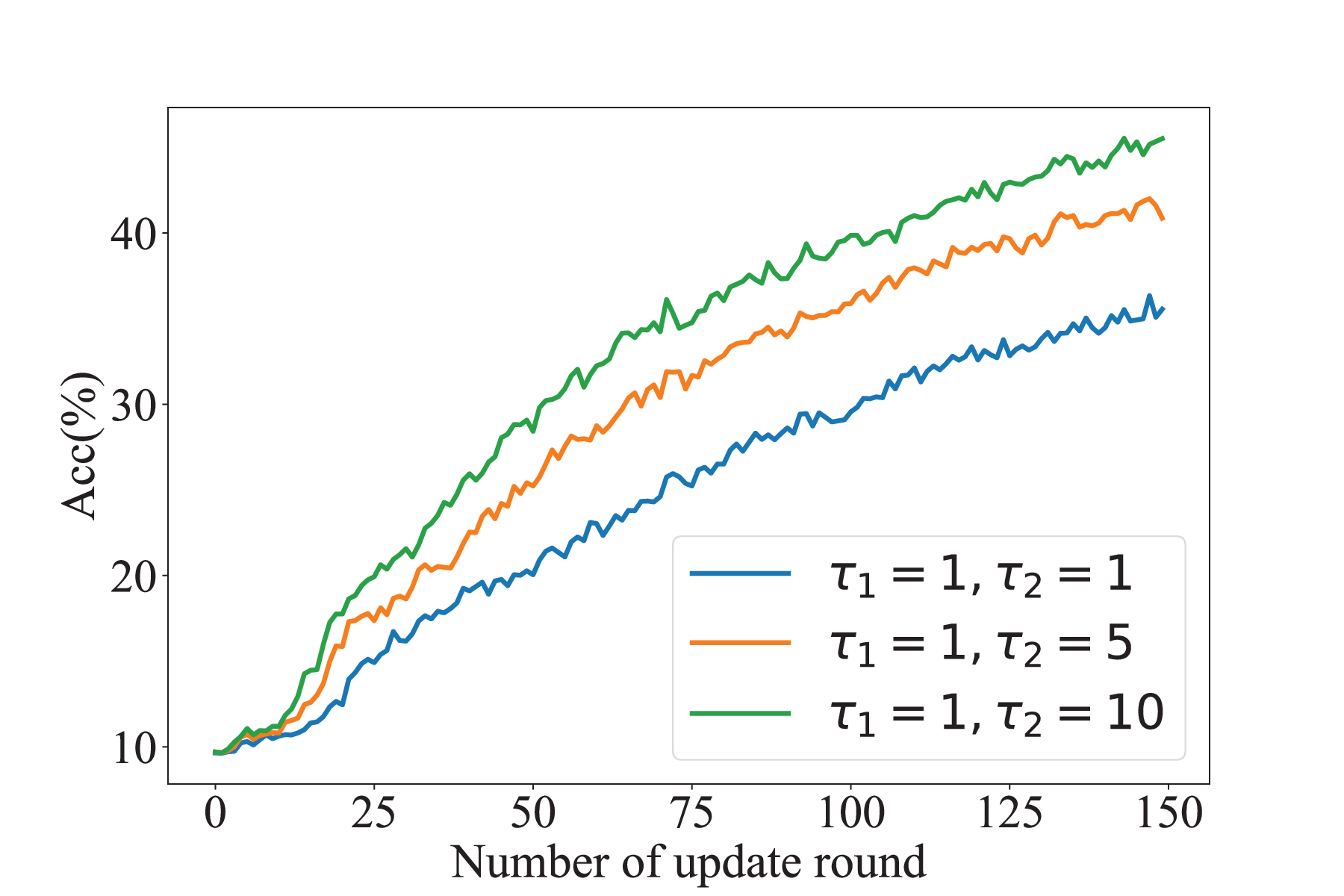}
		\label{fig3_2}}
	\subfloat[]{\includegraphics[width=2.4in]{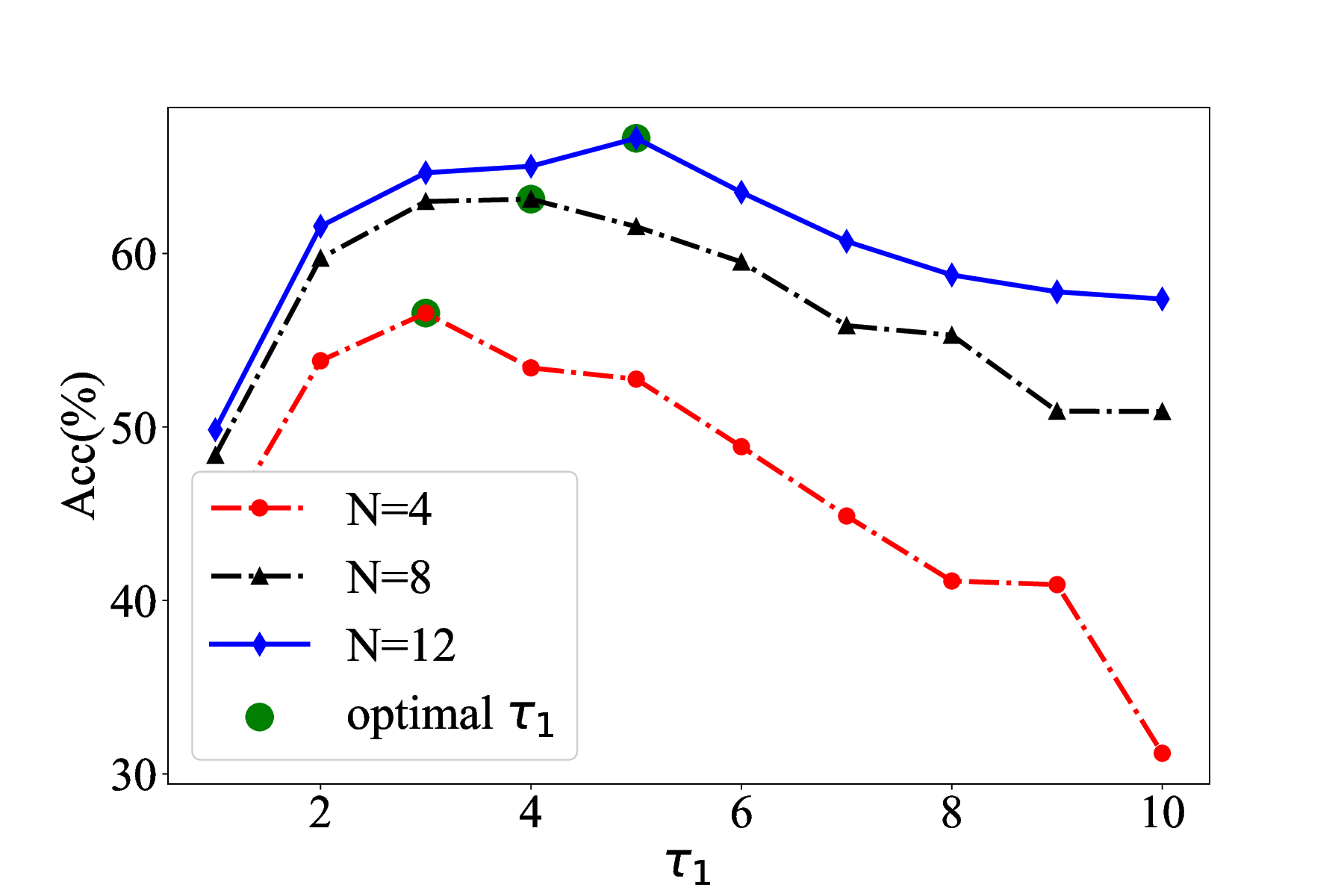}
		\label{fig3_3}}
	\caption{Impact of $\tau_{1}$ and $\tau_{2}$ on the accuracy with/without resource constraints. (a) Impact of $\tau_{1}$ on the accuracy without resource constraints. (b) Impact of $\tau_{2}$ on the accuracy without resource constraints. (c) Impact of $\tau_{1}$ and $N$ on the accuracy with resource constraints.}
\end{figure*}

In Fig. 3(a), we compare the model performance with 3 different $\tau_{1}$ (2, 5 and 10). The models are trained by DFL with 100 update rounds, which have $\tau_{1}$ rounds of local training and 10 rounds of communication in one update round. Fig. 3(a) shows that, if we ignore the resource constraints, less local training in one update round makes the model perform better. The cases $\tau_{1}=2$ and $\tau_{1}=5$ achieve over than 50 percent accuracy after 53 and 68 update rounds. Finally, these 3 cases achieve 58.99\%, 57.17\% and 48.53\% prediction accuracy after 100 update rounds, respectively. When $\tau_{1}$ is the same, the performance with different $\tau_{2}$ (1, 5 and 10) after 150 update rounds are shown in Fig. 3(b). The final prediction accuracy rates of the three cases are 35.08\%, 41.59\% and 45.34\%. 

\begin{figure}[!t]
	\vspace{-7mm}
	\centering
	\subfloat[]{\includegraphics[width=1.8in]{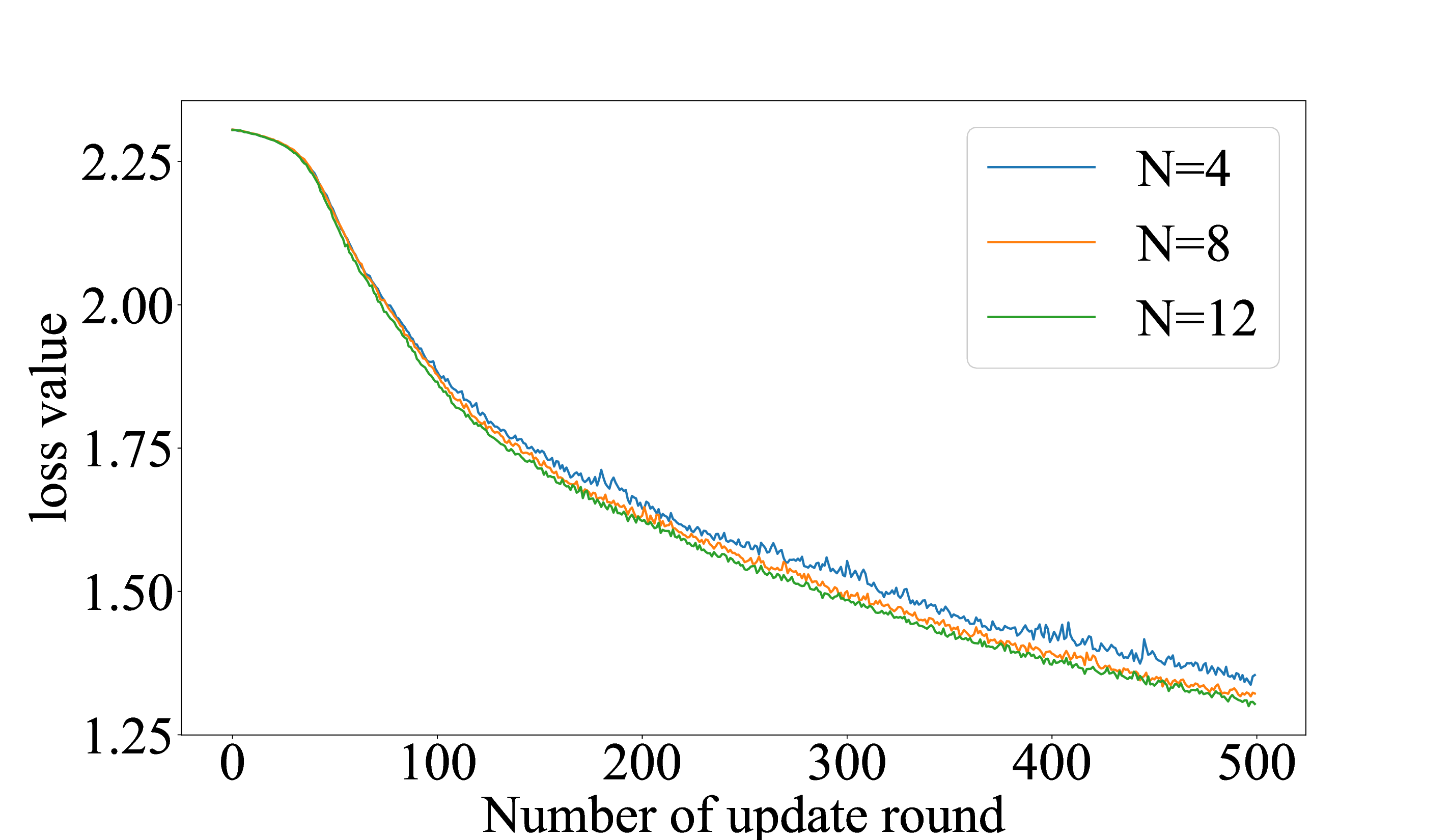}
		\label{fig7_1}}\hfil  
	\hspace{-0.26in}
	\subfloat[]{\includegraphics[width=1.8in]{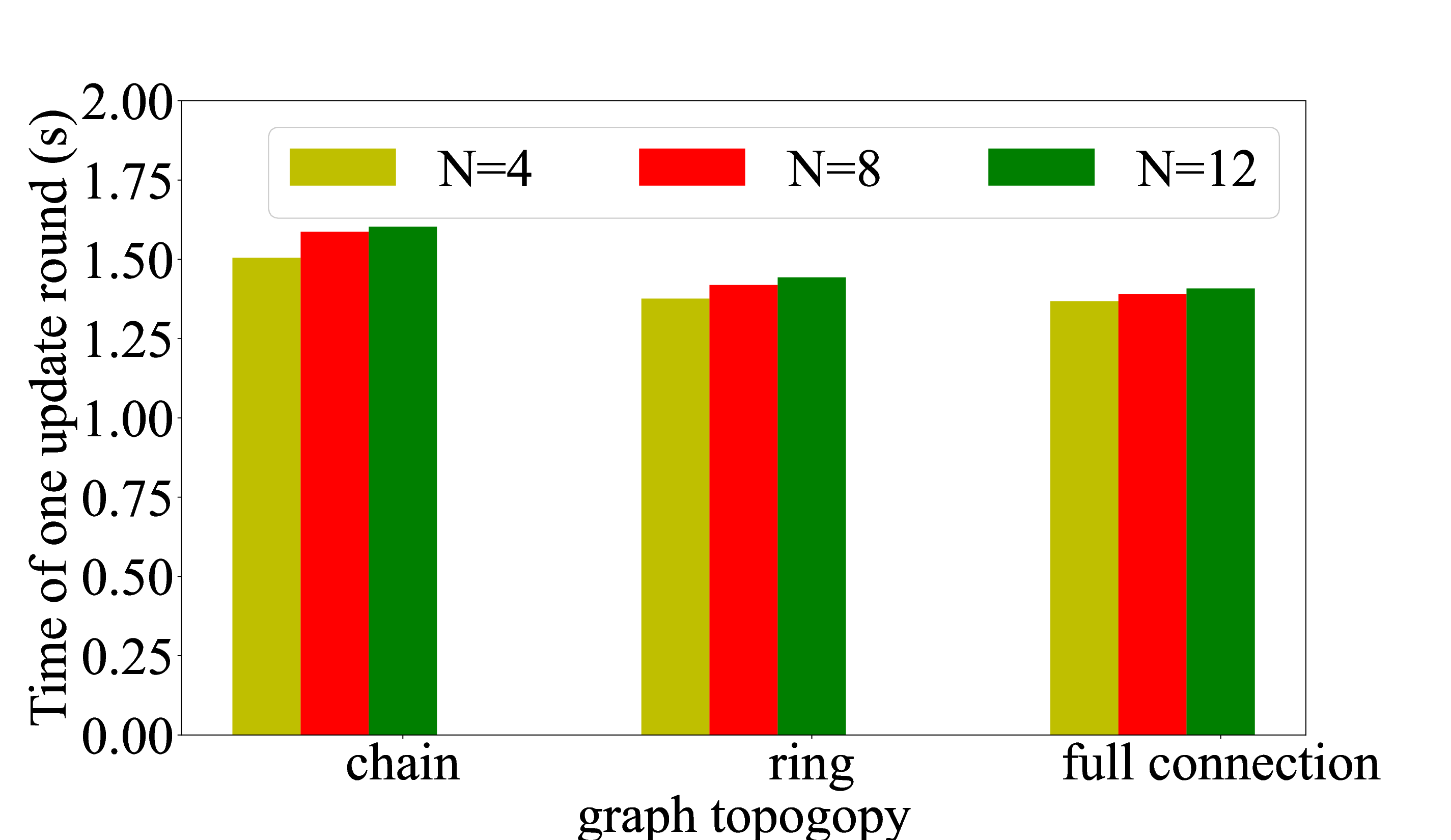}
		\label{fig7_2}}
	\caption{Impact on $N$ and graph topology. (a) Impact of $N$ on the convergence rate of DFL. (b) Impact of graph topology on the convergence rate of DFL.}
\end{figure}

These two simulation results are consistent with the discussion following Theorem 1. Without considering resource constraints, realizing information exchange between nodes through as much communication as possible helps the parameters of the DFL system to be closer to the global optimum. So it can achieve a better performance. On the contrary, more local training makes each node only reach its own local optimum, and has a larger error with the global optimum, which hurts the performance of the model.

\textit{2) DFL \textbf{with} resource constraints in digital transmission:} When DFL faces the cost constraints, more local training rounds means less communication. As the description in Corollary 2, if DFL has $T$ update rounds, we assume $(\tau_{1}+\tau_{2})T\leq T_{m}$. In this case, we let $T_{m}=10^{4}$ and $\tau_{2}=10$. The final prediction accuracy of DFL with different $\tau_{1}$ after $T_{m}$ is exhausted is compared. To make the simulation more comprehensive, we study 3 cases with different number of devices (e.g., $N=4,8,12$), which are shown in Fig. 3(c).

From Fig. 3(c), we know that, there is a trade-off between number of local training round and communication round. In the DFL with 4 devices, the best accuracy is 56.57\% when $\tau_{1}=3$. This value is $\tau_{1}=4$ and $\tau_{1}=5$ in cases $N=8$ and $N=12$. The corresponding accuracy are 63.13\% and 66.64\%, respectively. This result proves the discussion of \eqref{optt}, which says there is an optimal $\tau_{1}$ with limited resources, which makes the DFL outperform others under the same condition. Besides, this result also confirms Theorem 3 about the relationship between the number of devices and performance of DFL. More devices participating in DFL can help improve the performance of DFL-trained models. Furthermore, we observe that, the optimal $\tau_{1}$ becomes larger as the number of devices in DFL increases. In the case of resource constraints, since there is no guarantee that there will be enough communication for all devices to synchronize all information, local training after sharing some information can achieve a better performance. The simulation result implies that DFL with more devices require less communication to improve the performance, since it faces higher communication costs.

\textit{3) Convergence rate of DFL with different $N$ in digital transmission:} We compare the convergence rate with different number of devices in DFL. To ensure the fairness of the comparison, we set the same noise power ($\sigma_{n}=-30$dBm), learning rate ($\eta=0.01$) and communication error probability.

Fig. 4(a). shows the comparison result with $N=4,8,12$. It reflects the DFL with more devices has a higher convergence rate. In particular, when $N=4$, it takes 433 update rounds for the global loss function of DFL to converge to below 1.4, which are 385 and 373 when $N=8$ and $12$, respectively. This result is consistent with the analysis in Theorem 4. Besides, the difference in the convergence rate caused by different numbers of devices becomes more significant as the number of update rounds increases. This means that the convergence rate advantage brought by $N$ can be cumulative.

\textit{4) Impact of $\lambda_{2}$ in digital transmission:} In addition to the above influencing factors, the influence of the graph topology composed of the connection relationship of devices on the convergence rate is also studied. We selected three typical graph topologies, full connection, ring and chain, and the corresponding $\lambda_{2}$ are incremented sequentially.

In the DFL of this part, it has one local training round and one communication round in an update round. Thus, the convergence rate of DFL is evaluated by the time of one update round it uses. By analyzing the result of Fig. 4(b), we know that, a DFL with a higher connection density has a faster convergence rate. Likewise, when DFL corresponds to the same graph topology, more devices of DFL slows down the convergence, thereby verifying our analysis that the convergence upper bound decreases with $\lambda_{2}$ in Theorem 4.

\begin{figure}[!t]
	\vspace{-7mm}
	\centering
	\subfloat[]{\includegraphics[width=1.8in]{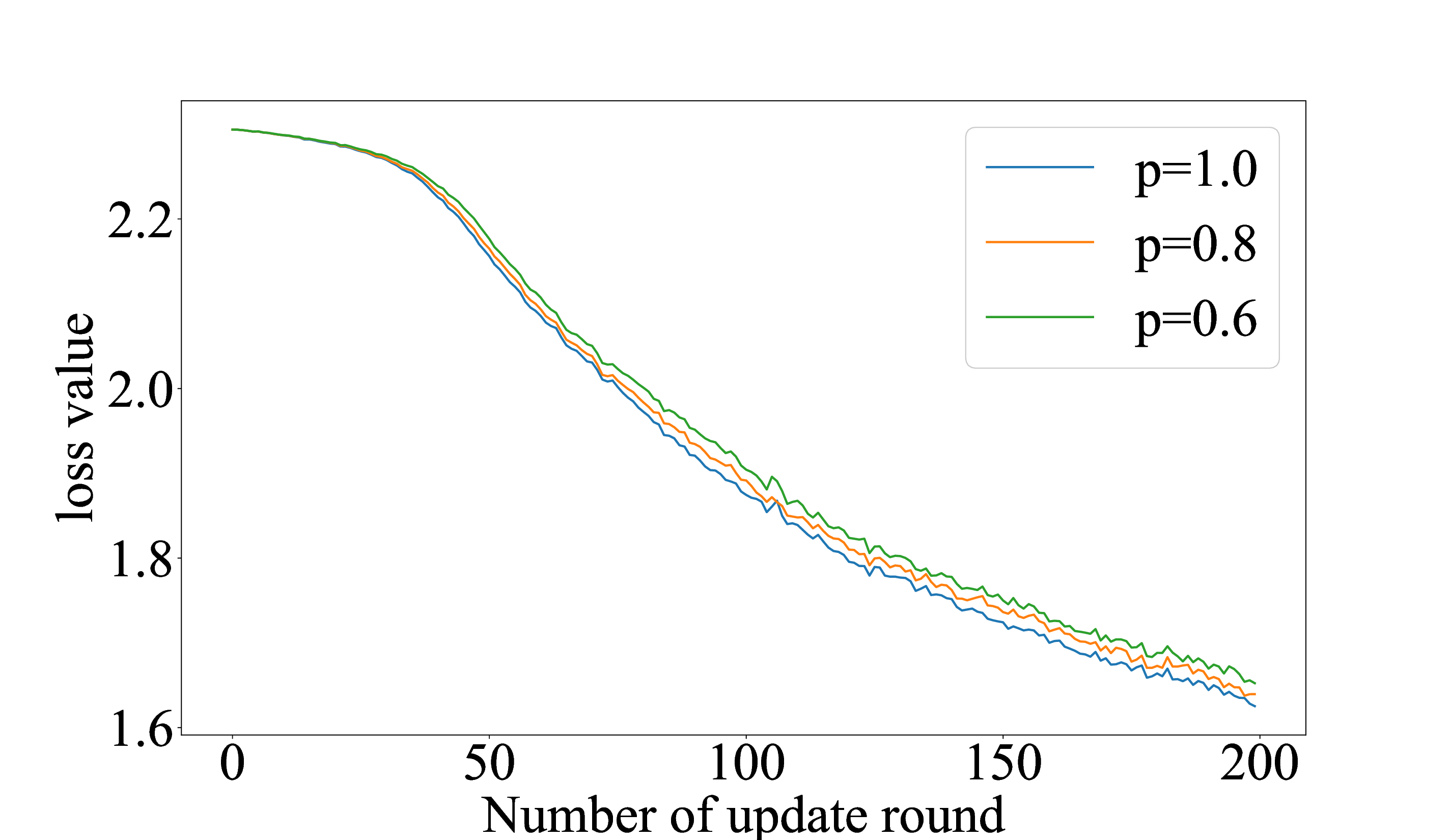}
		\label{fig8_1}}\hfil  
	\hspace{-0.3in}
	\subfloat[]{\includegraphics[width=1.8in]{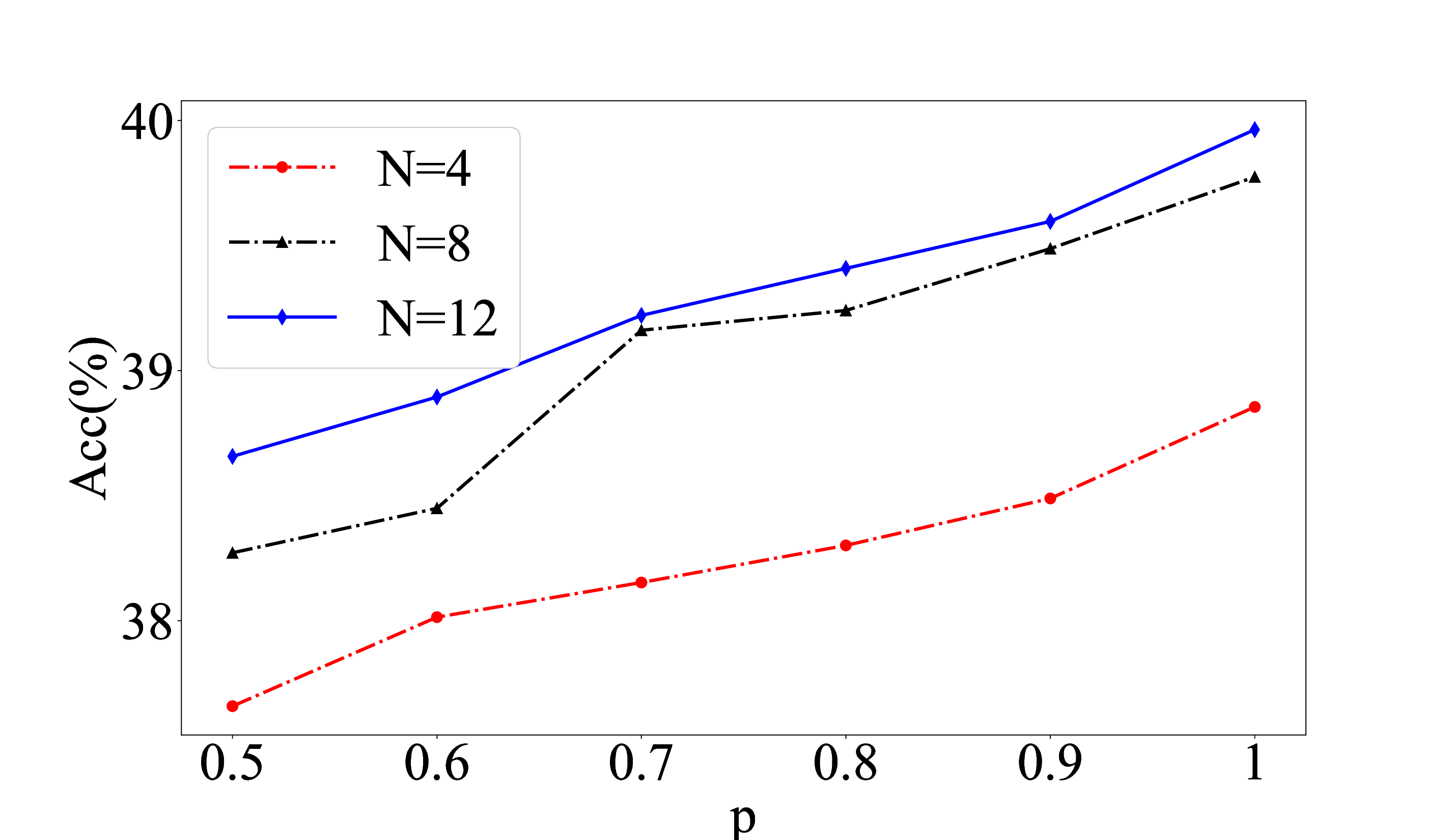}
		\label{fig8_2}}
	\caption{Impact on $p$. (a) Impact of $p$ on the convergence rate of DFL. (b) Impact of $p$ on the performance of DFL.}
\end{figure}

\textit{5) Impact of $p$ in digital transmission:} The probability of correction communication also affects the performance and convergence rate. In this part, we compare the convergence rate of DFL with 8 devices and different probability ($p=0.6,0.8,1$) at first. Then we compare their accuracy with different number of devices and correction probability to show the influence of $p$. The results are shown in Fig. 5(a) and 5(b). From Fig. 5(a), we know that, a higher $p$ means a faster converge. Specifically, in the case of $p=1$, the loss function of the DFL requires 120 rounds to be updated below 1.8. This number is 126 and 134 when $p=0.8$ and $p=0.6$. Similarly, according to Fig. 5(b), as $p$ increases, the performance of DFL is also increasing. In Fig. 9, we show the performance changes when $N=4,8$ and $12$ ($\eta=0.01$ and $T=200$ are fixed). When $p=0.5$, the corresponding accuracies are 37.65\%, 38.27\% and 38.66\%. As $p$ is improved, their accuracies become 38.85\%, 39.77\%, and 39.96\%.

Second, in DFL over fading channels, we discuss the effect of error accumulation level, which includes the channel gain, noise power and path loss, on the model convergence with different number of devices. Besides, by observing the convergence of DFL at different learning rates, we find that DFL has a maximum learning rate that guarantees the convergence. Therefore, we also compare the difference in the maximum acceptable learning rate for DFL with different numbers of devices under different noise powers.

\textit{1) Impact of $\kappa$ and $\sigma_{n}$ in analog transmission:}
In this part, we provide the accuracy of different error accumulation level to show the effect of channel gain, path loss and noise power on the performance after 200 update rounds.
\begin{figure*}[!t]
	\vspace{-1.5cm}
	\centering
	\subfloat[]{\includegraphics[width=2.4in]{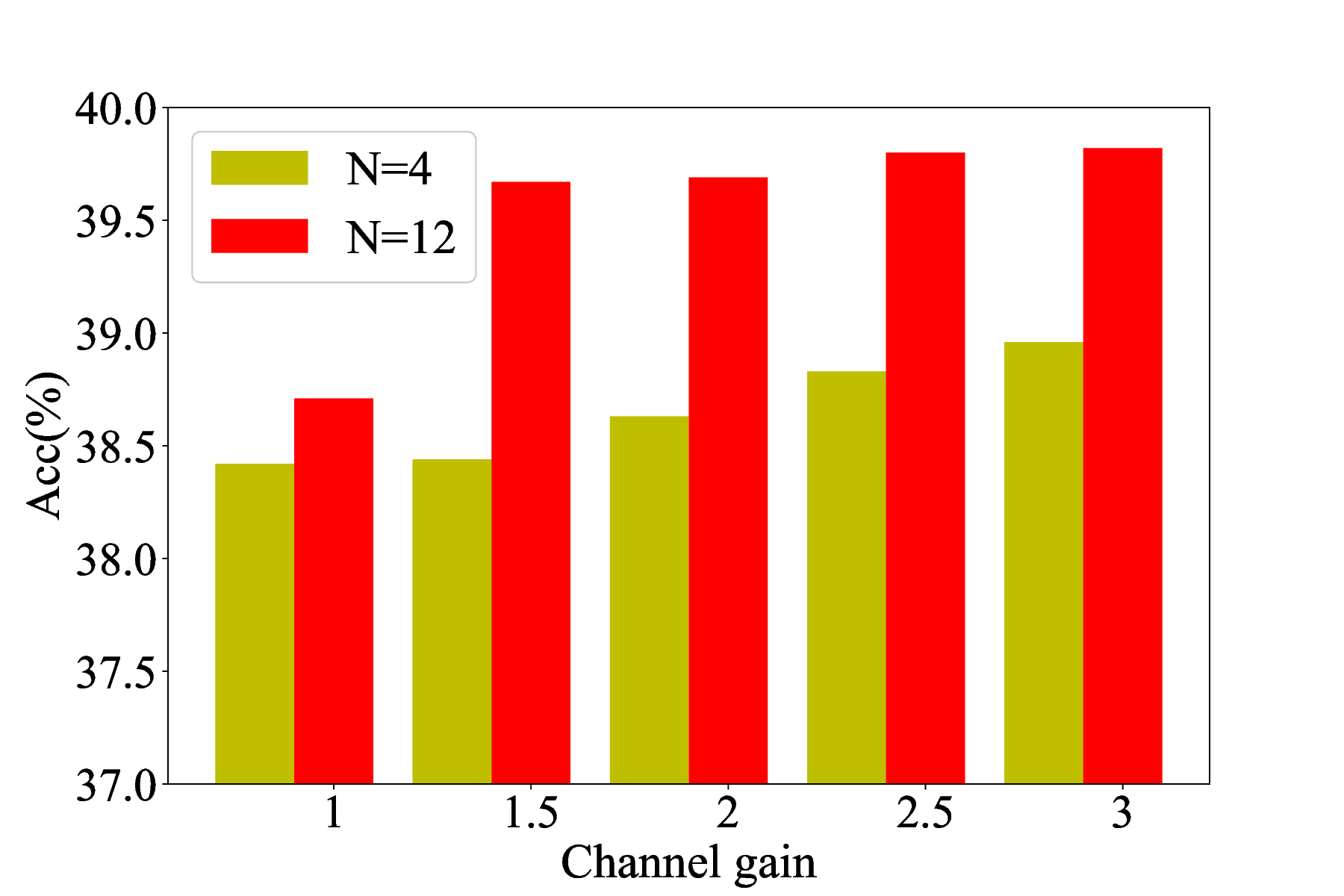}
		\label{fig10_1}}
	\subfloat[]{\includegraphics[width=2.4in]{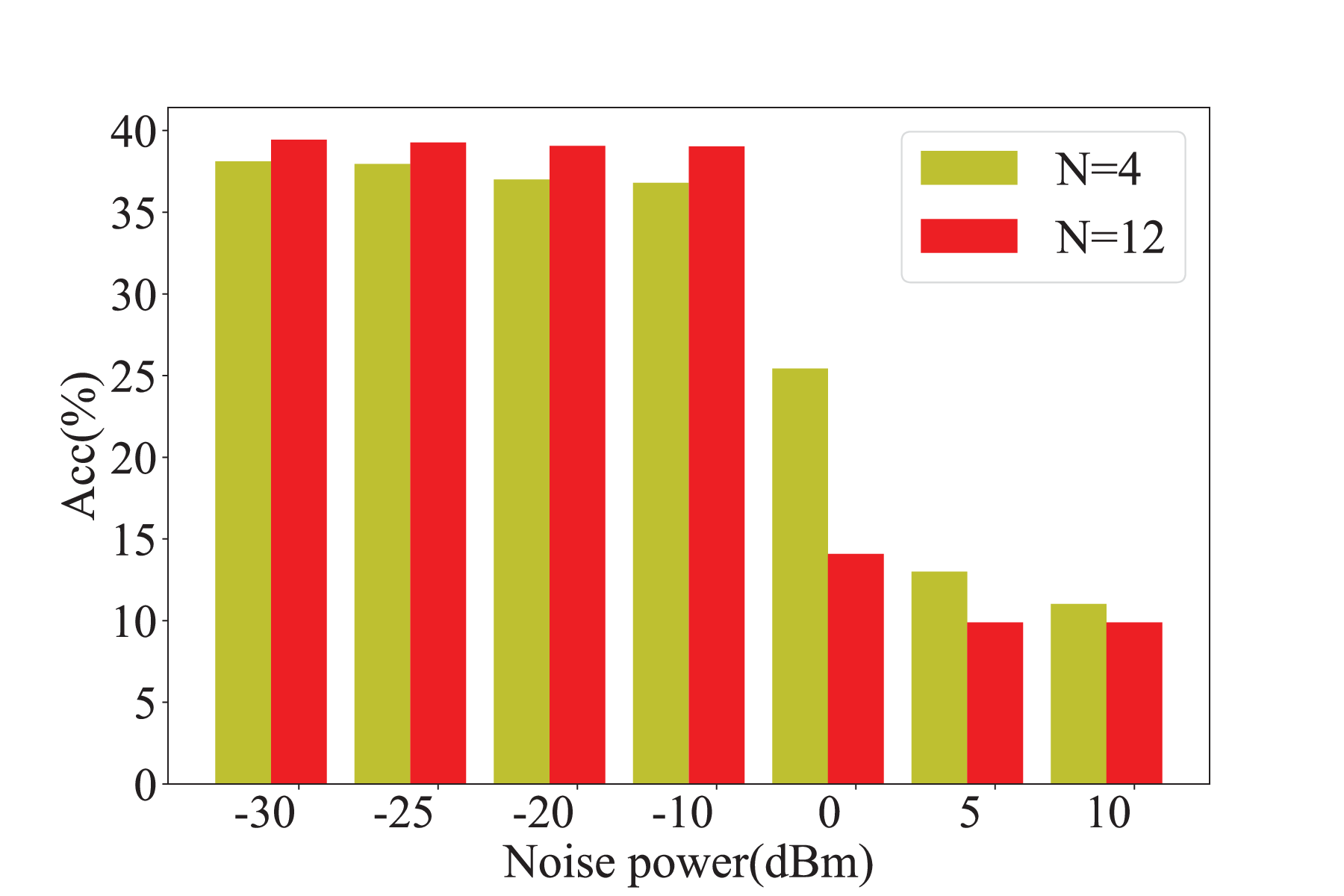}
		\label{fig10_2}}
	\subfloat[]{\includegraphics[width=2.4in]{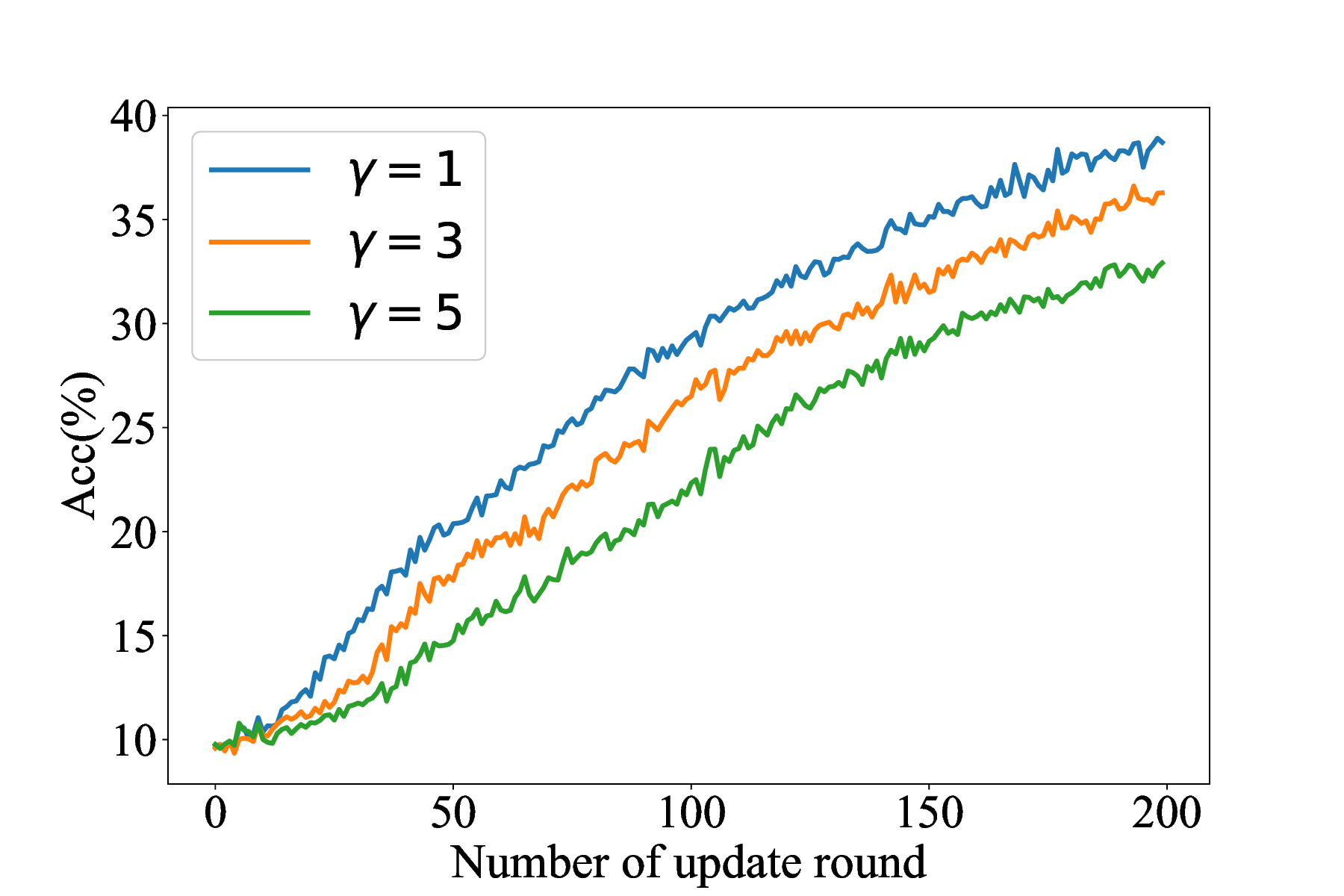}
		\label{fig10_3}}
	\caption{Impact of channel gain, noise power and path loss on the performance of DFL. (a) Impact of $N$ and $h$ on the performance of DFL with $\sigma_{n}=-20$dBm and $\gamma=1$. (b) Impact of $N$ and $\sigma_{n}$ on the performance of DFL with $h=1$ and $\gamma=1$. (c) Impact of $\gamma$ on the performance of DFL with $h=1$,  $\sigma_{n}=-20$dBm and $N=12$.}
\end{figure*}

\begin{figure}[!t]
	\vspace{-1cm}
	\centering
	\subfloat[]{\includegraphics[width=1.8in]{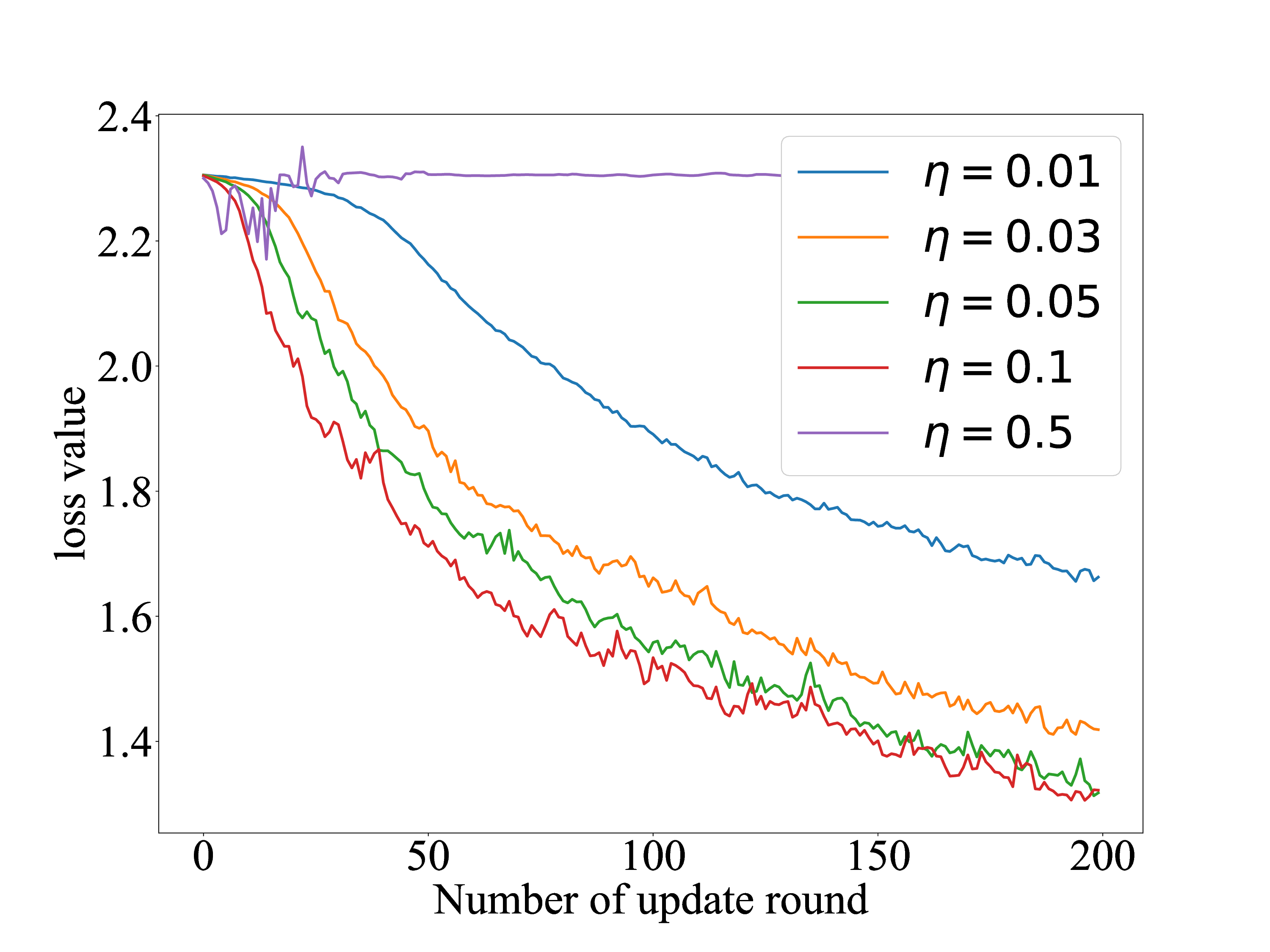}
		\label{fig12_1}}\hfil  
	\hspace{-0.3in}
	\subfloat[]{\includegraphics[width=1.8in]{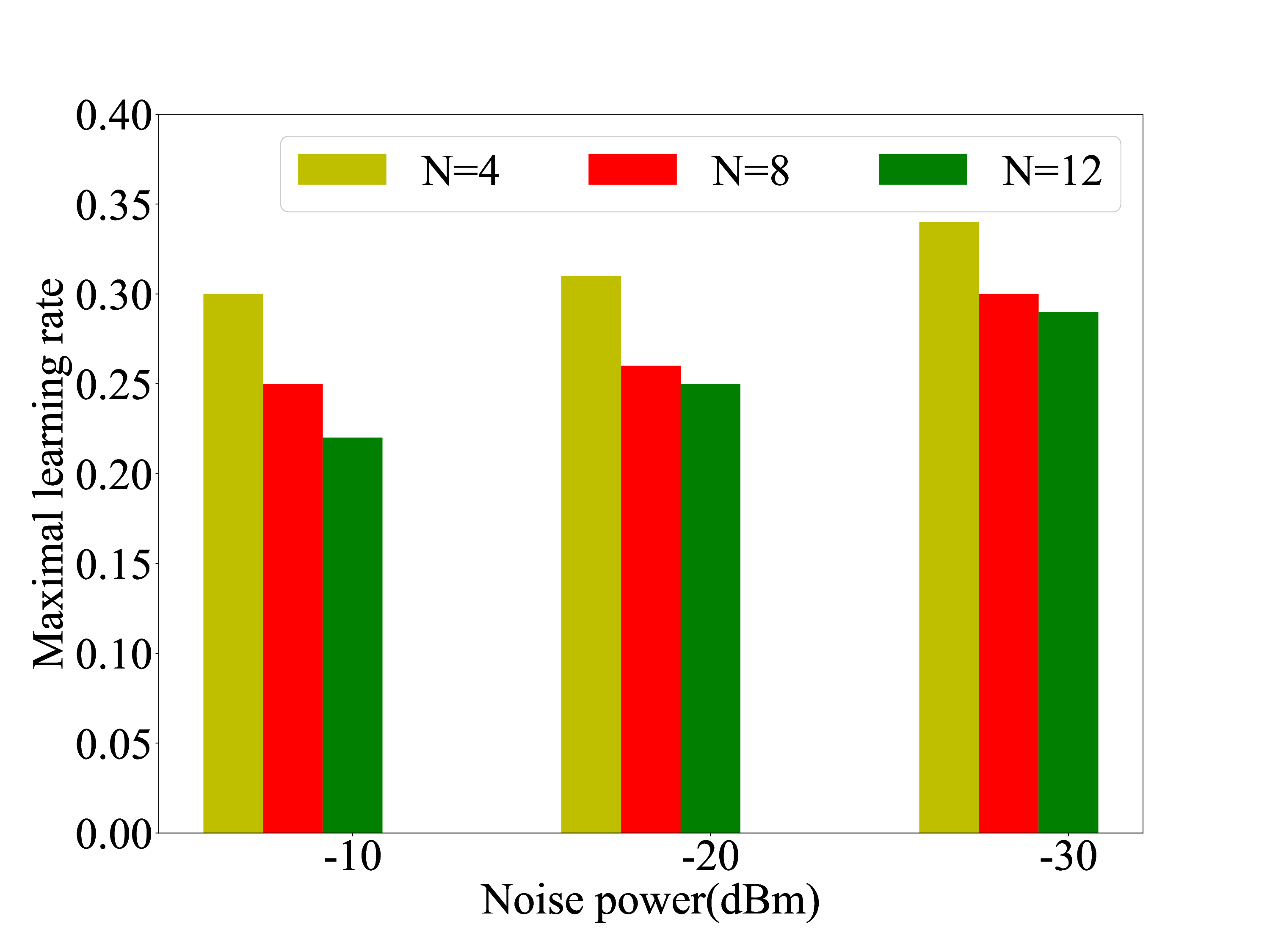}
		\label{fig12_2}}
	\caption{Impact on $\eta$. (a) Impact of $\eta$ on the convergence rate of DFL. (b) Impact of $N$ and $\sigma_{n}$ on $\eta$ of DFL.}
\end{figure}

The effect of error accumulation level on the performance of DFL are compared in Fig. 10, which includes the channel gain, noise power and path loss in 6(a), 6(b) and 6(c) respectively. Fig. 6(a) shows that, the prediction accuracy is improved with the channel gain increasing (a larger $|h|^{2}$) in both $N=4$ and $N=12$ cases. In addition, note that the impact of $|h|^{2}$ increasing on the model performance is more significant in the case with larger $N$. This observation implies that $\kappa$ and $N$ are coupled, which is consistent with the conclusion obtained in Theorem 3. From the Fig. 6(b), we know that, when the noise power is relatively small  ($-30$dBm to $-10$dBm), although greater power leads to worse accuracy, the performance of DFL with more devices participating is still better. When $N=12$, the accuracy changes from 39.44\% to 39.02\%, and this value range is from 38.11\% to 36.78\% in the case of $N=4$. Since there are 10 labels in CIFAR-10, and the accuracy is less than 10\%, we can consider that the DFL does not converge. When the noise power is 5dBm, the accuracy of DFL with 12 devices is 9.89\%. However, the accuracies of DFL with 4 devices are 12.99\% and 10.02\% in 5dBm and 10dBm noise. Besides, when the noise power is from 0dBm to 10dBm, the DFL with 4 devices performs better than that with 12 devices. It means that, the damage caused by the accumulation of noise is more significant in the DFL which has more devices. Such a system is also less tolerant to noise. This justifies our discussion and analysis for Corollary 3. Finally, the impact of path loss are shown is Fig. 6(c). With the $\gamma$ increasing, model performance becomes worse, which are 38.71\%, 36.28\% and 32.91\% after 200 update rounds, with $\gamma=1,3$ and $5$ respectively.

\begin{figure}[!t]
	\vspace{-1cm}
	\centering
	\subfloat[]{\includegraphics[width=1.8in]{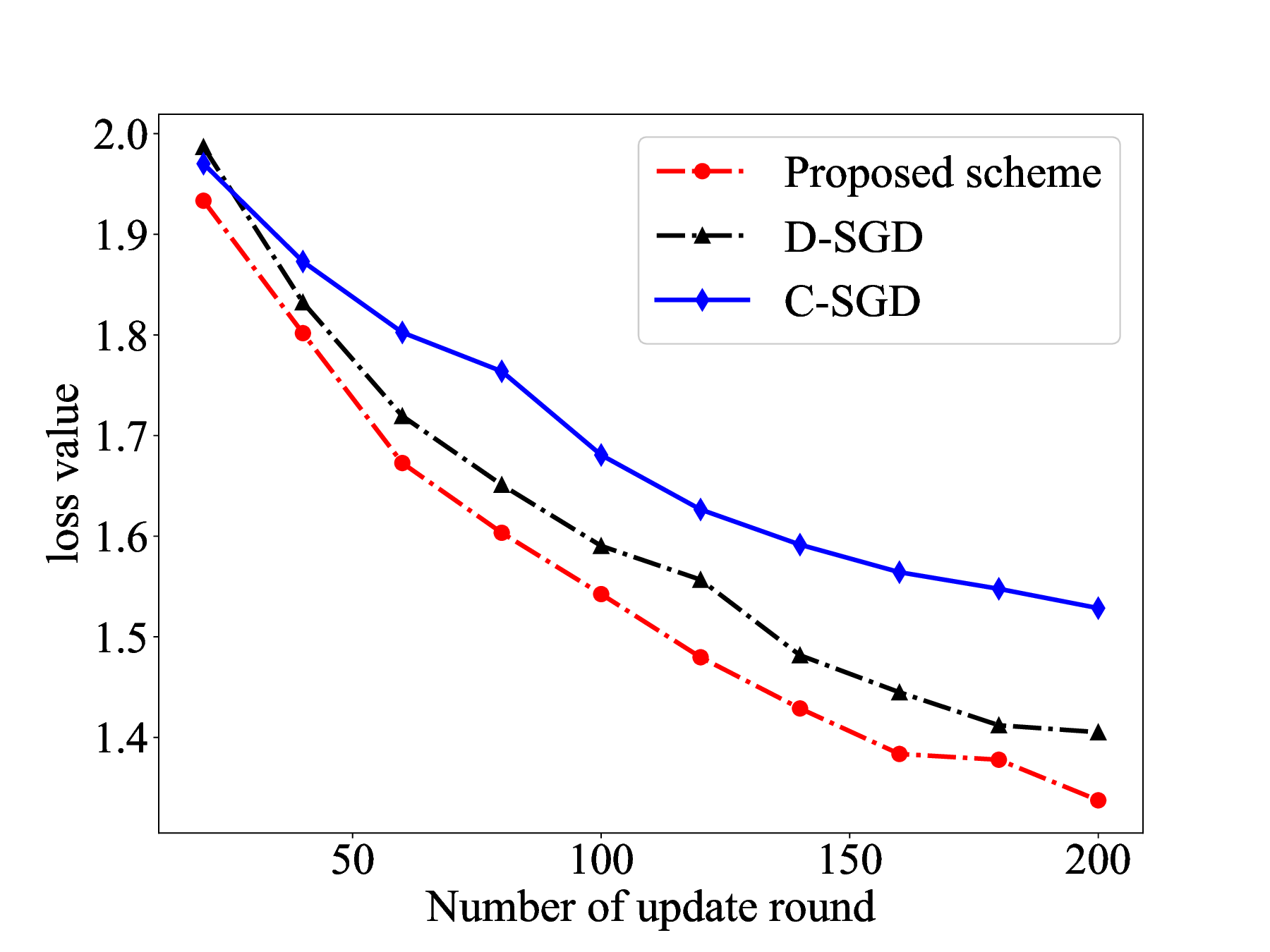}
		\label{fig2_1}}\hfil  
	\hspace{-0.3in}
	\subfloat[]{\includegraphics[width=1.8in]{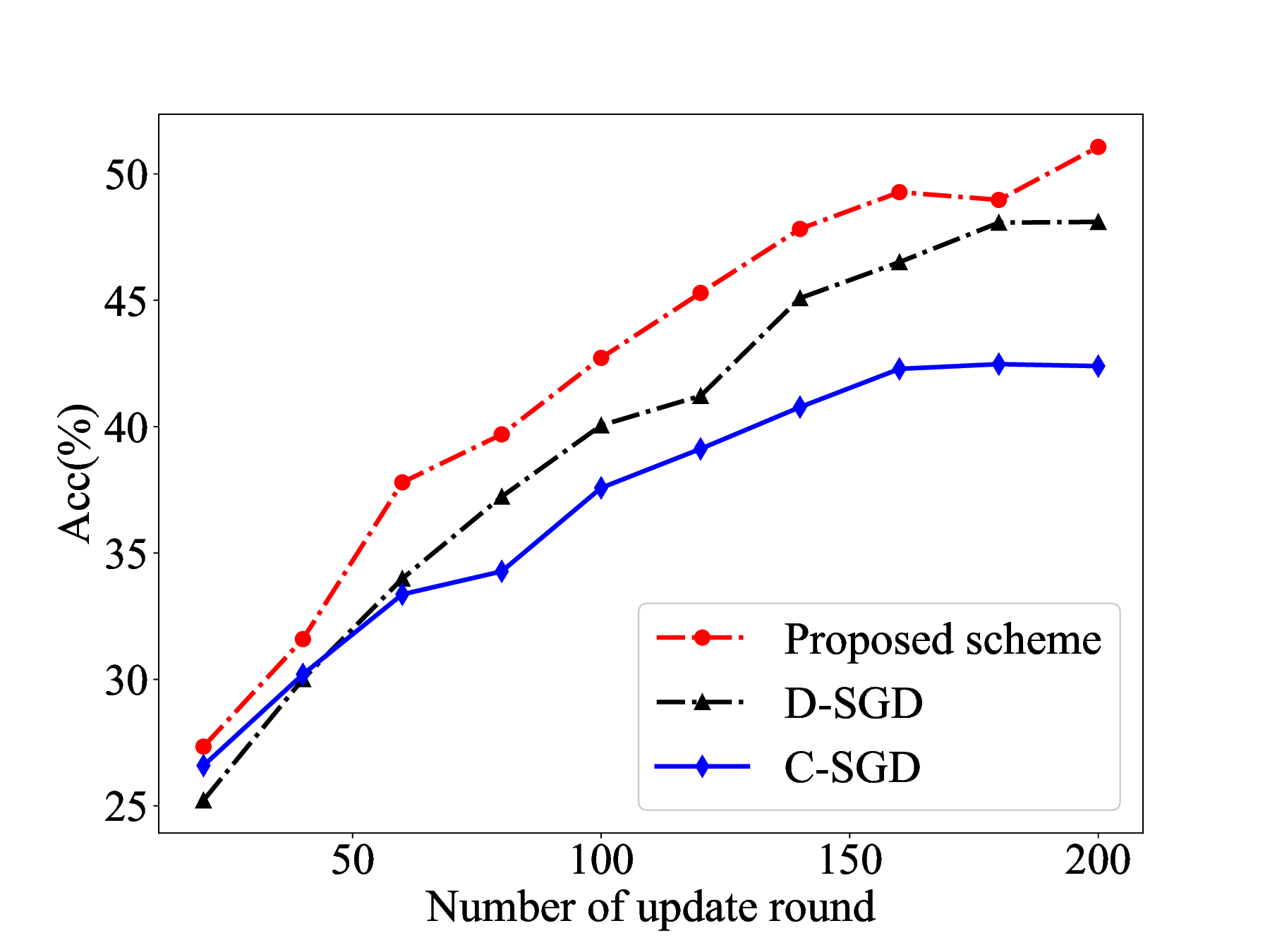}
		\label{fig2_2}}
	\caption{Comparison of benchmarks.}
\end{figure}

\textit{2) Impact of $\eta$ in analog transmission:}
In this part, we first compare the convergence rate with different learning rates. Then, combining the noise power and the number of devices, we study the maximum learning rate for DFL with different number of devices ($N=4,8,12$) to ensure the convergence for different noise powers ($-10$dBm, $-20$dBm and $-30$dBm). 

In Fig. 7(a), when $\eta$ is from 0.01 to 0.1, the convergence rate of DFL gradually increases. After 200 update rounds, 
when $\eta=0.1$ and $0.05$, the loss values are 1.318 and 1.322, respectively. In contrast, in the cases of $\eta=0.03$ and $0.01$, the loss values are 1.419 and 1.662, respectively. However, when $\eta=0.5$, DFL cannot converge. When $\eta$ is within a certain range, a larger $\eta$ brings a higher convergence rate for DFL. But too large $\eta$ causes DFL to fail to converge. Then, we investigate and analyze the effect of $\eta$ on the DFL convergence in Fig. 7(b). We compare the maximal learning rate to ensure the convergence of DFL with different power noises ($-10$dBm, $-20$dBm and $-30$dBm). To make our comparison more comprehensive, we consider the DFL with 4, 8 and 12 devices. In Fig. 7(b), for all 3 cases, the acceptable maximum $\eta$ increases with decreasing $\sigma_{n}$. Besides, under the same $\sigma_{n}$, DFL with fewer devices can tolerate a larger learning rate.

Finally, we compare the loss value and accuracy of the proposed recourse allocation scheme between computation and communication, and 2 learning strategies of DFL. In this comparison, D-SGD and C-SGD are set as the benchmarks, which are typical update schemes in DFL. D-SGD was proposed and studied in \cite{dfl1,can}. In this scheme, in one update round, all devices execute local training and communication once alternately. Different from D-SGD, the devices in C-SGD perform multiple local training and communication once in one update round. This scheme saves communication costs and improves efficiency by reducing the frequency of communication, which is proposed and discussed in \cite{dfl1,cgsd}. Thus, in this comparison, we set D-SGD with $\tau_{1}=\tau_{2}=1$ and C-SGD with $\tau_{1}=3,\tau_{2}=1$ as the benchmarks. The comparison results of loss values and accuracy are shown in Fig. 13. Besides, different from the previous part of simulations in this section, we consider a Vision Transformer (ViT) model to complete the classification task on CIFAR-10 in our comparison.

Fig. 8 shows that, the proposed update scheme of DFL with optimized $\tau_{1}$ and $\tau_{2}$ has a lower loss value and a higher accuracy than D-SGD and C-SGD. This is because the recourse allocation between local training and communication are not considered in these schemes. Thus, both of them are only communication once in one update round. However, in our previous simulations in this section, we have verified that under the same costs budget, appropriately increasing $\tau_{2}$ to achieve a balance between computation and communication can improve model performance. This balanced $\tau_{1}$ and $\tau_{2}$ can be obtain in our proposed scheme.

\section{Conclusions}

In this paper, we have studied the DFL under limited communication recourse constraints. First, we have proposed a comprehensive performance analysis of the DFL under both digital and analog communication approaches by deriving two upper bounds of the gap between the final loss function parameters and their global optimal values. These upper bounds reflect the effects of various system parameters, such as the number of devices and the number of local training and communication rounds, on the model performance, as well as the influence of communication factors, such as the probability of erroneous communication caused by package errors in digital transmission with UDP, the channel fading and noise in analog transmission, and total communication resources. Besides, we have also gain further insights by investigating the digital and analog transmission approaches separately. For digital transmission, we have analyzed the resource allocation scheme and convergence rate of DFL, derived the minimal probability of correct communication, and studied the impact of the density of connections between devices on convergence rate. For analog transmission, we have analyzed the impact of channel fading and noise, explored the influence of different variables on error accumulation caused by fading and noise, and derived the maximum error accumulation level that a DFL system can tolerate. Finally, simulation results have confirmed the effectiveness of the proposed analysis and discussions. Besides, we can obtain some system parameter settings to achieve a better performance. For example, it is desirable to consider fewer local updates in a DFL system with more devices, and reducing the number of devices participating in training in a communication scenario for a greater noise power.

{\appendices
	
\section{Proof of Theorem 1}

To prove Theorem 1, we introduce the following lemmas.

\textit{\textbf{Lemma 1}: An upper bound of the mathematical expectation of $\Vert\hat{\mathbf{Q}}_{T-1}\Vert_{\mathrm{F}}$ is given by}
\begin{equation}\small
	\begin{aligned}
		\label{L1}
		\mathbb{E}[\Vert\hat{\mathbf{Q}}_{T-1}\Vert_{\mathrm{F}}]\leq \sqrt{(\Vert\mathbf{Q}_{T-1}\Vert_{\mathrm{F}}^{2}-N)p^{2}+Np},
	\end{aligned}
\end{equation}
where $\hat{\mathbf{Q}}_{j}\triangleq\prod_{i=1}^{j\tau_{2}}\hat{\mathbf{P}}_{j}$.

\textit{Proof: See Appendix D. $\hfill\qedsymbol$}

\textit{\textbf{Lemma 2}: If $\mathbf{n}=[n_{ij}]_{M\times N}$ is a noise matrix with $n_{ij}\sim \mathcal{N}(0,\sigma^{2})$. Then, the upper bound of the mathematical expectation of $\Vert\mathbf{n}\Vert_{\mathrm{F}}$ is given by}
\begin{equation}\small
	\begin{aligned}
		\label{L2}
		\mathbb{E}[\Vert\mathbf{n}\Vert_{\mathrm{F}}]\leq \sqrt{MN\sigma^{2}}.
	\end{aligned}
\end{equation}

\textit{Proof: See Appendix E. $\hfill\qedsymbol$}

According to \eqref{sgd} and \eqref{boardcast11}, we have
	\begin{equation}\small
		\begin{aligned}
			\label{p11}
			\mathbf{w}_{2}&=\tilde{\mathbf{w}}_{2}\prod_{i=1}^{\tau_{2}}\hat{\mathbf{P}}_{i}
			=\Big(\mathbf{w}_{1}-\sum_{i=1}^{\tau_{1}}\eta_{t}\nabla F\big(\tilde{\mathbf{w}}_{1+\frac{i}{\tau_{1}}}\big)\Big)\prod_{i=1}^{\tau_{2}}\hat{\mathbf{P}}_{i}
			\\
			&=\mathbf{w}_{1}\hat{\mathbf{Q}}_{1}-\sum_{i=1}^{\tau_{1}}\eta_{t}\nabla F\big(\tilde{\mathbf{w}}_{1+\frac{i}{\tau_{1}}}\big)\hat{\mathbf{Q}}_{1}
			,
		\end{aligned}
	\end{equation}
and
	\begin{equation}\small
	\begin{aligned}
		\label{p12}
	    &\mathbf{w}_{3}=\mathbf{w}_{2}\hat{\mathbf{Q}}_{1}-\sum_{i=1}^{\tau_{1}}\eta_{t}\nabla F\big(\tilde{\mathbf{w}}_{2+\frac{i}{\tau_{1}}}\big)\hat{\mathbf{Q}}_{1}
	    \\
	    &=\mathbf{w}_{1}\hat{\mathbf{Q}}_{2}-\Big(\sum_{i=1}^{\tau_{1}}\eta_{t}\nabla F\big(\tilde{\mathbf{w}}_{1+\frac{i}{\tau_{1}}}\big)\hat{\mathbf{Q}}_{2} +\sum_{i=1}^{\tau_{1}}\eta_{t}\nabla F\big(\tilde{\mathbf{w}}_{2+\frac{i}{\tau_{1}}}\big)\hat{\mathbf{Q}}_{1}\Big)\\
	\end{aligned}
\end{equation} 
Because $\eta_{t}\triangleq\frac{\eta}{(t-1)^{2}}\geq\frac{\eta}{(T-1)^{2}}$, we have
	\begin{equation}\small
	\begin{aligned}
		\label{p13}
		\mathbf{w}_{T}&=\mathbf{w}_{1}\hat{\mathbf{Q}}_{T-1}-\sum_{j=1}^{T-1}\sum_{i=1}^{\tau_{1}}\eta_{t}\nabla F\big(\tilde{\mathbf{w}}_{j+\frac{i}{\tau_{1}}}\big)\hat{\mathbf{Q}}_{T-j}\\
		&\preceq\mathbf{w}_{1}\hat{\mathbf{Q}}_{T-1}-\frac{\eta}{(T-1)^{2}}\sum_{j=1}^{T-1}\sum_{i=1}^{\tau_{1}}\nabla F\big(\tilde{\mathbf{w}}_{j+\frac{i}{\tau_{1}}}\big)\hat{\mathbf{Q}}_{T-j}.\\
	\end{aligned}
\end{equation}
Then, according to (21) in \cite{error} and Corollary 2 in \cite{can}, the gradient of the DFL is bounded by
\begin{equation}\small
	\begin{aligned}
		\label{p14}
		\frac{1}{T}\sum_{i=1}^{T}\Vert \nabla F(\mathbf{w}_{i}) \Vert_{\mathrm{F}}^{2}\leq\frac{1}{\sqrt{NT}}+\frac{N}{T}.
	\end{aligned}
\end{equation}
By using \eqref{con} and triangle inequality, we obtain
	\begin{equation}\small
	\begin{aligned}
		\label{p15}
		&\mathbb{E}[\Vert \mathbf{w}_{T}-\mathbf{w^{*}} \Vert_{\mathrm{F}}]\leq\underbrace{\mathbb{E}[\Vert \mathbf{w}_{1}\hat{\mathbf{Q}}_{T-1} \Vert_{\mathrm{F}}]}_{B_{1}}\\
		&+\underbrace{\mathbb{E}\Big[\Big\Vert \frac{\eta}{(T-1)^{2}}\sum_{j=1}^{T-1}\sum_{i=1}^{\tau_{1}}\nabla F\big(\tilde{\mathbf{w}}_{j+\frac{i}{\tau_{1}}}\big)\hat{\mathbf{Q}}_{T-j} \Big\Vert_{\mathrm{F}}\Big]}_{B_{2}}+\Vert \mathbf{w^{*}} \Vert_{\mathrm{F}}.
	\end{aligned}
\end{equation}
Based on Lemma 1, we have
\begin{equation}\small
	\begin{aligned}
		\label{p16}
		B_{1}&{\leq} \Vert \mathbf{w}_{1}\Vert_{\mathrm{2}}\mathbb{E}[\Vert\hat{\mathbf{Q}}_{T-1} \Vert_{\mathrm{F}}]{\leq}\Vert \mathbf{w}_{1}\Vert_{\mathrm{2}}\sqrt{(\Vert\mathbf{Q}_{T-1}\Vert_{\mathrm{F}}^{2}{-}N)p^{2}{+}Np},
	\end{aligned}
\end{equation}
and
\begin{equation}\small
	\begin{aligned}
		\label{p17}
		&B_{2}\leq \mathbb{E}\Big[\frac{\eta}{(T-1)^{2}}\sum_{i=1}^{\tau_{1}}\Big\Vert \sum_{j=1}^{T-1}\nabla F\big(\tilde{\mathbf{w}}_{j+\frac{i}{\tau_{1}}}\big)\hat{\mathbf{Q}}_{T-j} \Big\Vert_{\mathrm{F}}\Big]\\
		&\leq \frac{\eta}{(T-1)^{2}}\sum_{i=1}^{\tau_{1}}\mathbb{E}\Big[\Big\Vert \sum_{j=1}^{T-1}\nabla F\big(\tilde{\mathbf{w}}_{j+\frac{i}{\tau_{1}}}\big)\sum_{j=1}^{T-1}\hat{\mathbf{Q}}_{T-j} \Big\Vert_{\mathrm{F}}\Big]\\
		&\leq \frac{\eta}{(T-1)^{2}}\sum_{i=1}^{\tau_{1}}\Big\Vert \sum_{j=1}^{T-1}\nabla F\big(\tilde{\mathbf{w}}_{j+\frac{i}{\tau_{1}}}\big)\Big\Vert_{\mathrm{F}}\mathbb{E}\Big[\Big\Vert\sum_{j=1}^{T-1}\hat{\mathbf{Q}}_{j} \Big\Vert_{\mathrm{F}}\Big]\\
		&\leq \frac{\eta}{T-1}\sum_{i=1}^{\tau_{1}}\Big(\frac{1}{T-1}\sum_{j=1}^{T-1}\Big\Vert \nabla F\big(\tilde{\mathbf{w}}_{j+\frac{i}{\tau_{1}}}\big)\Big\Vert_{\mathrm{F}}\Big)\mathbb{E}\Big[\sum_{j=1}^{T-1}\Vert\hat{\mathbf{Q}}_{j} \Vert_{\mathrm{F}}\Big]\\
		&\leq \eta\tau_{1}\Big(\frac{1}{\sqrt{N(T-1)}}+\frac{N}{T-1}\Big)\sqrt{(\Vert\mathbf{Q}_{T-1}\Vert_{\mathrm{F}}-N)p^{2}+Np}.
	\end{aligned}
\end{equation}
Then, according to Assumption 1, we complete the proof.

\section{Proof of Theorem 2}
To prove Theorem 3, we introduce the following lemma.

\textit{\textbf{Lemma 3}: An upper bound of the mathematical expectation of $\Vert\hat{\mathbf{Q}}_{T-1}-\mathbf{I}\Vert_{\mathrm{F}}$ is given by}
\begin{equation}\small
	\begin{aligned}
		\label{L3}
		\mathbb{E}[\Vert\hat{\mathbf{Q}}_{T-1}-\mathbf{I}\Vert_{\mathrm{F}}]\leq \sqrt{(\Vert\mathbf{Q}_{T-1}\Vert_{\mathrm{F}}^{2}-N)p^{2}+Np+N}.
	\end{aligned}
\end{equation}

\textit{Proof: See Appendix F. $\hfill\qedsymbol$}

First, by using the Markov inequality, we have
\begin{equation}\small
	\label{p41}
	\mathrm{Pr}\Big(\frac{\Vert\mathbf{w}_{i,T}-\mathbf{w}_{i}^{*}\Vert_{\mathrm{2}}}{\Vert\mathbf{w}_{i,1}\Vert_{\mathrm{2}}+\delta}\geq\epsilon\Big)\leq \frac{\epsilon^{-1}\mathbb{E}[\Vert\mathbf{w}_{i,T}-\mathbf{w}_{i}^{*}\Vert_{\mathrm{2}}]}{\Vert\mathbf{w}_{i,1}\Vert_{\mathrm{2}}+\delta},
\end{equation}
Then, using the triangle inequality, we have
	\begin{equation}\small
	\begin{aligned}
		\label{p42}
		&\mathbb{E}[\Vert\mathbf{w}_{i,T}-\mathbf{w}_{i}^{*}\Vert_{\mathrm{2}}]=\mathbb{E}\Big[\Big\Vert \mathbf{w}_{i,1}\hat{\mathbf{Q}}_{T-1}-\mathbf{w}_{i}^{*}\\
		&~~~~~~-\frac{\eta}{(T-1)^{\frac{3}{2}}}\sum_{j=1}^{T-1}\sum_{k=1}^{\tau_{1}}\nabla F_{i}\big(\tilde{\mathbf{w}}_{j+\frac{k}{\tau_{1}}}\big)\hat{\mathbf{Q}}_{T-j}\Big\Vert_{\mathrm{2}}\Big]\\
		&\leq\mathbb{E}\Big[\Big\Vert \mathbf{w}_{i,1}\hat{\mathbf{Q}}_{T-1}-\mathbf{w}_{i}^{*}\Big\Vert_{\mathrm{2}}\Big]\\
		&~~~~~~+\mathbb{E}\Big[\Big\Vert\frac{\eta}{(T-1)^{\frac{3}{2}}}\sum_{j=1}^{T-1}\sum_{k=1}^{\tau_{1}}\nabla F_{i}\big(\tilde{\mathbf{w}}_{j+\frac{k}{\tau_{1}}}\big)\hat{\mathbf{Q}}_{T-j}\Big\Vert_{\mathrm{2}}\Big]
	\end{aligned}
\end{equation}
Since
\begin{equation}\small
	\label{p43}
    \frac{A+B}{x+y}=\frac{A}{x+y}+\frac{B}{x+y}<\frac{A}{x}+\frac{B}{y},
\end{equation}
by combing \eqref{p41}-\eqref{p43}, we obtain
\begin{equation}\small
	\begin{aligned}
	\label{p44}
	&\mathrm{Pr}\Big(\frac{\Vert\mathbf{w}_{i,T}-\mathbf{w}_{i}^{*}\Vert_{\mathrm{2}}}{\Vert\mathbf{w}_{i,1}\Vert_{\mathrm{2}}+\delta}\geq\epsilon\Big)\leq\underbrace{\frac{\epsilon^{-1}\mathbb{E}[\Vert\mathbf{w}_{i,1}\hat{\mathbf{Q}}_{T-1}-\mathbf{w}_{i}^{*}\Vert_{\mathrm{2}}]}{\Vert\mathbf{w}_{i,1}\Vert_{\mathrm{2}}}}_{B'_{1}}\\
	&+\underbrace{\frac{\epsilon^{-1}}{\delta}\mathbb{E}\Big[\Big\Vert\frac{\eta}{(T-1)^{\frac{3}{2}}}\sum_{j=1}^{T-1}\sum_{k=1}^{\tau_{1}}\nabla F_{i}\big(\tilde{\mathbf{w}}_{j+\frac{k}{\tau_{1}}}\big)\hat{\mathbf{Q}}_{T-j}\Big\Vert_{\mathrm{2}}\Big]}_{B'_{2}}.
	\end{aligned}
\end{equation}
Then, based on Lemma 3 and triangle inequality, we have
\begin{equation}\small
	\begin{aligned}
		\label{p45}
		B'_{1}&=\frac{\epsilon^{-1}\mathbb{E}[\Vert \mathbf{w}_{i,1}\hat{\mathbf{Q}}_{T-1}-\mathbf{w}_{i,1}+\mathbf{w}_{i,1}-\mathbf{w}_{i}^{*}\Vert_{\mathrm{2}}]}{\Vert\mathbf{w}_{i,1}\Vert_{\mathrm{2}}}\\
		&\leq\epsilon^{-1}\Big(\frac{\mathbb{E}[\Vert \mathbf{w}_{i,1}\hat{\mathbf{Q}}_{T-1}-\mathbf{w}_{i,1}\Vert_{\mathrm{2}}]}{\Vert\mathbf{w}_{i,1}\Vert_{\mathrm{2}}}+H\Big)\\
		&\leq\epsilon^{-1}\big(\mathbb{E}[\Vert \hat{\mathbf{Q}}_{T-1}-\mathbf{I}\Vert_{\mathrm{2}}]+H\big)\\
		&\leq\epsilon^{-1}\Big(\sqrt{(\Vert\mathbf{Q}_{T-1}\Vert_{\mathrm{F}}^{2}-N)p^{2}+Np+N}+H\Big).
	\end{aligned}
\end{equation}
Since $0\leq\lambda_{n}\leq\lambda_{n-1}\leq\cdots\lambda_{2}\leq\lambda_{1}=1$, according to the Rayleigh quotient theorem, we have
\begin{equation}\small
	\begin{aligned}
		\label{p46}
		x^{T}\mathbb{E}[P^{T}P]x\leq\lambda_{1}(P^{T}P)\Vert x\Vert_{\mathrm{2}}^{2}\leq\frac{1}{1-\lambda_{2}(P^{T}P)}\Vert x\Vert_{\mathrm{2}}^{2}.
	\end{aligned}
\end{equation}
Thus, $B'_{2}$ is bounded by
\begin{equation}\small
	\begin{aligned}
		\label{p47}
		&B'_{2}\leq \frac{\epsilon^{-1}}{\delta}\Big(\frac{\eta}{(T-1)^{\frac{3}{2}}}\mathbb{E}\Big[\Big\Vert\sum_{j=1}^{T-1}\sum_{k=1}^{\tau_{1}}\nabla F_{i}\big(\tilde{\mathbf{w}}_{j+\frac{k}{\tau_{1}}}\big)\hat{\mathbf{Q}}_{T-j}\Big\Vert_{\mathrm{2}}\Big]  \Big)\\
		&\leq \frac{\epsilon^{-1}}{\delta}\Bigg(\frac{\eta}{(T-1)^{\frac{3}{2}}}\sqrt{\mathbb{E}\Big[\Big\Vert\sum_{j=1}^{T-1}\sum_{k=1}^{\tau_{1}}\nabla F_{i}\big(\tilde{\mathbf{w}}_{j+\frac{k}{\tau_{1}}}\big)\hat{\mathbf{Q}}_{T-j}\Big\Vert_{\mathrm{2}}^{2}\Big]  }\Bigg)\\
		&\leq \frac{\epsilon^{-1}}{\delta}\Bigg(\frac{\eta}{(T-1)^{\frac{3}{2}}}\sqrt{\sum_{j=1}^{T-1}\mathbb{E}\Big[\Big\Vert\sum_{k=1}^{\tau_{1}}\nabla F_{i}\big(\tilde{\mathbf{w}}_{j+\frac{k}{\tau_{1}}}\big)\hat{\mathbf{Q}}_{T-j}\Big\Vert_{\mathrm{2}}^{2}\Big]  }\Bigg)\\
		&\leq \frac{\epsilon^{-1}}{\delta}\Bigg(\frac{\eta}{(T-1)^{\frac{3}{2}}}\\
	&~~\sqrt{\sum_{j=1}^{T-1}\mathbb{E}\Big[\Big\Vert\sum_{k=1}^{\tau_{1}}\nabla F_{i}\big(\tilde{\mathbf{w}}_{j+\frac{k}{\tau_{1}}}\big)\Big\Vert_{\mathrm{2}}^{2}\Big(1-\lambda_{2}(\hat{\mathbf{Q}}_{T-j}^{T}\hat{\mathbf{Q}}_{T-j})\Big)^{-1}\Big]}\Bigg)\\
		&\leq \frac{\epsilon^{-1}}{\delta}\Bigg(\frac{\eta}{(T-1)^{\frac{3}{2}}}\sqrt{\sum_{j=1}^{T-1}\tau_{1}^{2}G^{2}\Big(1-\lambda_{2}(\hat{\mathbf{Q}}_{T-j}^{T}\hat{\mathbf{Q}}_{T-j})\Big)^{-1}}\Bigg)\\
		&= 
		\frac{\epsilon^{-1}}{\delta}\Bigg(\frac{\eta\tau_{1}G}{(T-1)^{\frac{3}{2}}}\cdot\bar{\beta}\sqrt{T-1}\Bigg)=\frac{\epsilon^{-1}}{\delta}\cdot\frac{\eta\tau_{1}G\bar{\beta}}{T-1}.
	\end{aligned}
\end{equation}
By combing \eqref{p44}, \eqref{p45} and \eqref{p47}, we have
\begin{equation}\small
	\begin{aligned}
		\label{p410}
		&\mathrm{Pr}\Big(\frac{\Vert\mathbf{w}_{i,T}-\mathbf{w}_{i}^{*}\Vert_{\mathrm{2}}}{\Vert\mathbf{w}_{i,1}\Vert_{\mathrm{2}}+\delta}\geq\epsilon\Big)\\
		&\leq\sqrt{(\Vert\mathbf{Q}_{T-1}\Vert_{\mathrm{F}}^{2}-N)p^{2}+Np+N}+H+\frac{\eta\tau_{1}G\bar{\beta}}{\delta(T-1)}\leq \epsilon^{2}.
	\end{aligned}
\end{equation}
After some basic manipulations, we complete the proof.

\section{Proof of Theorem 3}

Since the fixed mixing weights matrix $\mathbf{P}$ can be seen as the random matrix $\hat{\mathbf{P}}_{i}$ with $p=1$, similar to \eqref{p11}-\eqref{p13}, we have
	\begin{equation}\small
	\begin{aligned}
		\label{p31}
		\hat{\mathbf{w}}_{T}
		&\preceq\hat{\mathbf{w}}_{1}\hat{\mathbf{Q}}_{T-1}-\frac{\eta}{(T-1)^{2}}\sum_{j=1}^{T-1}\sum_{i=1}^{\tau_{1}}\nabla F\big(\tilde{\mathbf{w}}_{j+\frac{i}{\tau_{1}}}\big)\hat{\mathbf{Q}}_{T-j}\\
		&\quad\quad +\sum_{i=1}^{\tau_{2}}\tilde{\mathbf{n}}_{i}\prod_{i=1}^{\tau_{2}+i-1}\hat{\mathbf{P}}_{i}\Big(\mathbf{I}+\sum_{i=1}^{T-2}\hat{\mathbf{Q}}_{i}\Big).\\
	\end{aligned}
\end{equation}
Thus, according to \eqref{p14}, we obtain
	\begin{equation}\small
	\begin{aligned}
		\label{p32}
		&\mathbb{E}[\Vert\hat{\mathbf{w}}_{T}-\mathbf{w^{*}} \Vert_{\mathrm{F}}]\leq B_{1}+B_{2}\\
		&+\underbrace{\mathbb{E}\Big[\Big\Vert\sum_{i=1}^{\tau_{2}}\tilde{\mathbf{n}}_{i}\prod_{i=1}^{\tau_{2}+i-1}\hat{\mathbf{P}}_{i}\Big(\mathbf{I}+\sum_{i=1}^{T-2}\hat{\mathbf{Q}}_{i}\Big)\Big\Vert_{\mathrm{F}}\Big]}_{B_{3}}+\Vert\mathbf{w^{*}} \Vert_{\mathrm{F}},
	\end{aligned}
\end{equation}
where $B_{1}$ and $B_{2}$ are defined in \eqref{p15}. Since $\prod_{i=1}^{\tau_{2}+i-1}\hat{\mathbf{P}}_{i}$ and $\mathbf{Q}_{i}$ are also Markov matrices, the noise matrix multiplied by them is still a noise matrix. Thus, based on Lemma 2, we have
\begin{equation}\small
	\begin{aligned}
			\label{p33}
			B_{3}&=\mathbb{E}\Big[\Big\Vert\sum_{i=1}^{\tau_{2}}\tilde{\mathbf{n}}_{i}\prod_{i=1}^{\tau_{2}+i-1}\hat{\mathbf{P}}_{i}\Big(\mathbf{I}+\sum_{i=1}^{T-2}\hat{\mathbf{Q}}_{i}\Big)\Big\Vert_{\mathrm{F}}\Big]\\
			&=\mathbb{E}\Big[\Big\Vert (T-1)\sum_{i=1}^{\tau_{2}}\mathbf{n}_{i}' \Big\Vert_{\mathrm{F}}\Big]\leq (T-1)\sqrt{{\tau_{2}mN\kappa^{2}\sigma_{n}^{2}}}.
		\end{aligned}
\end{equation}
Then, according to Assumption 1, we complete the proof.

\section{Proof of Corollary 5}

Since $A_{3}(\sigma_{n})$ is an increasing function of $\sigma_{n}$, we have $A_{3}(\sigma_{n})\geq A_{3}(0)$ directly. For $A_{1}(p)$ and $A_{2}(p,T)$. Furthermore, since $(\Vert\mathbf{Q}_{T-1}\Vert_{\mathrm{F}}^{2}-N)p^{2}+Np$ is a quadratic function of $p$, and when
\begin{equation}
	\begin{aligned}
		\label{p21}
		p\geq\frac{N}{2(N-\Vert\mathbf{Q}_{T-1}\Vert_{\mathrm{F}}^{2})},
	\end{aligned}
\end{equation}
$A_{1}(p)$ and $A_{2}(p,T)$ are decreasing functions of $p$. When $N$ is large enough, we notice that, $\frac{N}{2(N-\Vert\mathbf{Q}_{T-1}\Vert_{\mathrm{F}}^{2})}\approx\frac{1}{2}$. Since the value range of $p$ is $(0.5,1]$, it follows that $A_{1}(p)$ and $A_{2}(p,T)$ are larger than $A_{1}(1)$ and $A_{2}(1,T)$. This completes the proof.

\section{Proof of Lemma 1}
Since $\mathbb{E}[X]^{2}=\mathbb{E}[X^{2}]-Var[X]\leq\mathbb{E}[X^{2}]$, we have
\begin{equation}\small
	\begin{aligned}
		\label{L11}
		&\mathbb{E}[\Vert\hat{\mathbf{Q}}_{T-1}\Vert_{\mathrm{F}}]\leq \sqrt{\mathbb{E}[\Vert\hat{\mathbf{Q}}_{T-1}\Vert_{\mathrm{F}}^{2}]}\\
		&=\sqrt{\mathbb{E}[\hat{q}_{11}^{2}+\cdots+\hat{q}_{NN}^{2}]}=\sqrt{\mathbb{E}[\hat{q}_{11}^{2}]+\cdots+\mathbb{E}[\hat{q}_{NN}^{2}]}\\
		&=\sqrt{\mathbb{E}[\hat{q}_{11}]^{2}+Var[\hat{q}_{11}]+\cdots+\mathbb{E}[\hat{q}_{NN}]^{2}+Var[\hat{q}_{NN}]}\\
		&=\sqrt{p^{2}(q_{11}^{2}+\cdots+q_{NN}^{2})+p(1-p)(q_{11}+\cdots+q_{NN})}\\
		&=\sqrt{\Vert\mathbf{Q}_{T-1}\Vert_{\mathrm{F}}^{2}p^{2}+Np(1-p)}=\sqrt{(\Vert\mathbf{Q}_{T-1}\Vert_{\mathrm{F}}^{2}-N)p^{2}+Np}
	\end{aligned}
\end{equation}
This completes the proof.

\section{Proof of Lemma 2}
Since $\mathbb{E}[X]^{2}=\mathbb{E}[X^{2}]-Var[X]\leq\mathbb{E}[X^{2}]$, we have
\begin{equation}\small
	\begin{aligned}
		\label{L21}
		&\mathbb{E}[\Vert\mathbf{n}\Vert_{\mathrm{F}}]\leq \sqrt{\mathbb{E}[\Vert\mathbf{n}\Vert_{\mathrm{F}}^{2}]}=\sqrt{\mathbb{E}[n_{11}^{2}+\cdots+n_{MN}^{2}]}\\
		&=\sqrt{\mathbb{E}[n_{11}^{2}]+\cdots+\mathbb{E}[n_{MN}^{2}]}\\
		&=\sqrt{Var[\hat{q}_{11}]+\cdots+Var[\hat{q}_{NN}]}=\sqrt{MN\sigma^{2}}
	\end{aligned}
\end{equation}
This completes the proof.

\section{Proof of Lemma 3}
Since $\mathbb{E}[X]^{2}=\mathbb{E}[X^{2}]-Var[X]\leq\mathbb{E}[X^{2}]$, we have
\begin{equation}\small
	\begin{aligned}
		\label{L31}
		\mathbb{E}&[\Vert\hat{\mathbf{Q}}_{T-1}-\mathbf{I}\Vert_{\mathrm{F}}]\leq \sqrt{\mathbb{E}[\Vert\hat{\mathbf{Q}}_{T-1}-\mathbf{I}\Vert_{\mathrm{F}}^{2}]}\\
		&=\sqrt{\mathbb{E}[(\hat{q}_{11}-1)^{2}+\hat{q}_{12}^{2}+\cdots+(\hat{q}_{NN}-1)^{2}]}\\
		&=\sqrt{\mathbb{E}[\hat{q}_{11}^{2}+\cdots+\hat{q}_{NN}^{2}]-2\mathbb{E}[\hat{q}_{11}+\hat{q}_{22}\cdots+\hat{q}_{NN}]+N}\\
		&=\sqrt{(\Vert\mathbf{Q}_{T-1}\Vert_{\mathrm{F}}^{2}-N)p^{2}+Np+N}
	\end{aligned}
\end{equation}
This completes the proof.

}
 
\bibliographystyle{ieeetr}
\bibliography{ref}

\vfill

\end{document}